\documentclass[sigconf, nonacm, pdfa]{acmart}

\usepackage{colorprofiles}
\usepackage[a-2b,mathxmp]{pdfx}[2018/12/22]
\hypersetup{pdfstartview=}

\usepackage{float}
\usepackage{booktabs}
\usepackage{amsmath}
\usepackage{amsthm}
\usepackage{graphicx}
\usepackage{balance} 
\usepackage{needspace} 
\usepackage{booktabs}
\usepackage{algorithm}
\usepackage{algpseudocode}
\usepackage{diagbox}
\usepackage{tabularx}
\usepackage{verbatim}
\usepackage{balance}

\usepackage[colorinlistoftodos]{todonotes}
\newtheorem{definition}{Definition}

\begin{document}
\title{CLaP - State Detection from Time Series}

%%
%% The "author" command and its associated commands are used to define the authors and their affiliations

\author{Arik Ermshaus}
\affiliation{%
  \institution{Humboldt-Universit\"at zu Berlin}
  \city{Berlin}
  \country{Germany}
}
\email{ermshaua@informatik.hu-berlin.de}

\author{Patrick Sch\"afer}
\affiliation{%
  \institution{Humboldt-Universit\"at zu Berlin}
  \city{Berlin}
  \country{Germany}
}
\email{patrick.schaefer@hu-berlin.de}

\author{Ulf Leser}
\affiliation{%
  \institution{Humboldt-Universit\"at zu Berlin}
  \city{Berlin}
  \country{Germany}
}
\email{leser@informatik.hu-berlin.de}

%%
%% The abstract is a short summary of the work to be presented in the
%% article.

\begin{abstract}
The ever-growing amount of sensor data from machines, smart devices, and the environment leads to an abundance of high-resolution, unannotated time series (TS). These recordings encode recognizable properties of latent states and transitions from physical phenomena that can be modelled as abstract processes. The unsupervised localization and identification of these states and their transitions is the task of time series state detection (TSSD). Current TSSD algorithms employ classical unsupervised learning techniques, to infer state membership directly from feature space. This limits their predictive power, compared to supervised learning methods, which can exploit additional label information. We introduce CLaP, a new, highly accurate and efficient algorithm for TSSD. It leverages the predictive power of time series classification for TSSD in an unsupervised setting by applying novel self-supervision techniques to detect whether data segments emerge from the same state. To this end, CLaP cross-validates a classifier with segment-labelled subsequences to quantify confusion between segments. It merges labels from segments with high confusion, representing the same latent state, if this leads to an increase in overall classification quality. We conducted an experimental evaluation using 405 TS from five benchmarks and found CLaP to be significantly more precise in detecting states than six state-of-the-art competitors. It achieves the best accuracy-runtime tradeoff and is scalable to large TS. We provide a Python implementation of CLaP, which can be deployed in TS analysis workflows.
\end{abstract}

% TSSD is an important data mining problem that preprocesses long, unannotated time series data, providing important input to data management and processing, advanced TS analytics such as anomaly detection, or decision-support for domain experts. 

% as distinct parts separated by sudden shifts in the signal.

\settopmatter{printfolios=true}
\maketitle

\sloppy

\section{Introduction} \label{sec:introduction}
% motivate why clustering is not enough

% For instance, acceleration recordings from smart devices can differentiate between a person walking and jogging~\cite{Lara2013ASO}, or a machine requiring maintenance or not~\cite{Kanawaday2017MachineLF}. Similarly, electrocardiograms reveal if a patient suffers from arrhythmias~\cite{Moody2001TheIO}, and seismometer measurements capture the onset of earthquakes~\cite{Woollam2022SeisBenchATF}. 

% The data management literature distinguishes between different types of subsequences in TS data. For instance, \emph{anomalies} are unusual, rare, or novel subsequences~\cite{Schmidl2022AnomalyDI}, \emph{motifs} are recurring structures~\cite{Schfer2022MotifletsS}, and \emph{segments} are prolonged, distinct parts of TS that capture process states~\cite{Ermshaus2024ClaSS}. 

% and their automatic localization and identification are crucial steps in many data analysis workflows and decision-support systems.

The study and analysis of long-running biological, human-controlled, or physical processes, such as human activities or industrial manufacturing, is of great interest in their respective domains. The field of human activity recognition, for instance, aims to detect falls in the elderly~\cite{Yin2008SensorBasedAH}. To achieve this, human motions are conceptualized as an abstract process of distinct activities that transition from one to another. Data acquisition and analysis workflows are employed to detect activities from mobile sensors and report falls. Similarly, in industrial manufacturing, CNC machines are monitored by pre-installed IoT devices to detect tool wear~\cite{Tran2022MachineLA}. The condition of a tool can be abstracted as either normal or faulty, and wear detection is realized by recognising the transitions in-between. From a computational perspective, such phenomena can be modelled as abstract processes with discrete states and pairwise transitions between them. The core property of these processes is that the states are distinct and can be observed and measured over time by instrumentation. 

%, and exhibit a fixed set of possible transitions. 
% IoT devices are used to capture the current, vibration, or sound of the system, and signal processing techniques are applied to identify the tool’s condition. 

Recent advances in the digitalization of measurement devices have greatly facilitated the acquisition of biological and physical process data. Sensors, such as those in smartphones, industrial machinery, or earth observation  stations, continuously record high-frequency real-valued observational data, termed \emph{time series} (TS)~\cite{Lara2013ASO,Kanawaday2017MachineLF,Woollam2022SeisBenchATF}. State changes in the observed processes lead to variations in the recorded signals, which form the basis for analysis. Sensor data capture the unique characteristics of states as subsequences, distinguishable by either their shape or statistical properties. These distinct parts of TS are commonly called \emph{segments}, and each contains the recognizable properties of the observed process state. The transitions between segments, also called \emph{change points} (CPs), mark the time points in a recording, at which the observed process changes state. The identification of segments, CPs, and state labels forms the foundation for activity recognition~\cite{ErmshausHumanAS}, health assessment~\cite{Kemp2000AnalysisOA}, or condition monitoring in IoT~\cite{Tran2022MachineLA}. 

% For example, the first part of a human activity recording might capture a person walking, followed by jogging, and then walking again. The transitions between walking and jogging, also called change points (CPs), mark the time points in the recording at which the observed process changes state. 

Formally, the task of recovering the sequence of discrete state labels from a TS of observations is called time series state detection (TSSD). It acts as a complex, unsupervised preprocessing step between data collection and TS knowledge discovery~\cite{Wang2024UnsupervisedTS}. TSSD annotates each data point in a TS with a label that corresponds to a state in the data-generating process. Consecutive stretches of the same label mark segments, while different neighbouring labels indicate CPs. TSSD requires the analysis of observations to identify signal shifts, which are assumed to result from state changes in the observed process. This creates consecutive segments of data points that are homogeneous within themselves, yet sufficiently distinct from their neighbours. These segments are further compared to each other to assign equal labels to those sharing the same state and unequal labels to those representing different states. State labels from segments are then propagated to individual data points to form the annotation, a state sequence. Contrary to supervised problems, such as TS classification (TSC), the labels in TSSD (e.g. 0,1,2) are discovered by an algorithm, abstract, and primarily used to differentiate observations based on state affiliation. They can be mapped to semantic labels (e.g. walk, jog, run) when appropriate domain knowledge is available~\cite{Ahad2021IoTSA}.

% It is commonly approached as an unsupervised data mining problem~\cite{Wang2023Time2StateAU}, where labels are machine-generated and primarily used to recognize recurring states and differentiate distinct ones. 

% This sequence encodes important information, such as segment and state duration as well as observed transitions. 

% Besides its own use, TSSD also enables the application of advanced analytics, such as anomaly detection~\cite{Paparrizos2022TSBUADAE}, to create normal models per state and predict outliers, for instance, to locate machine failures~\cite{skab}.

\begin{figure}[t]
    \includegraphics[width=1.0\columnwidth]{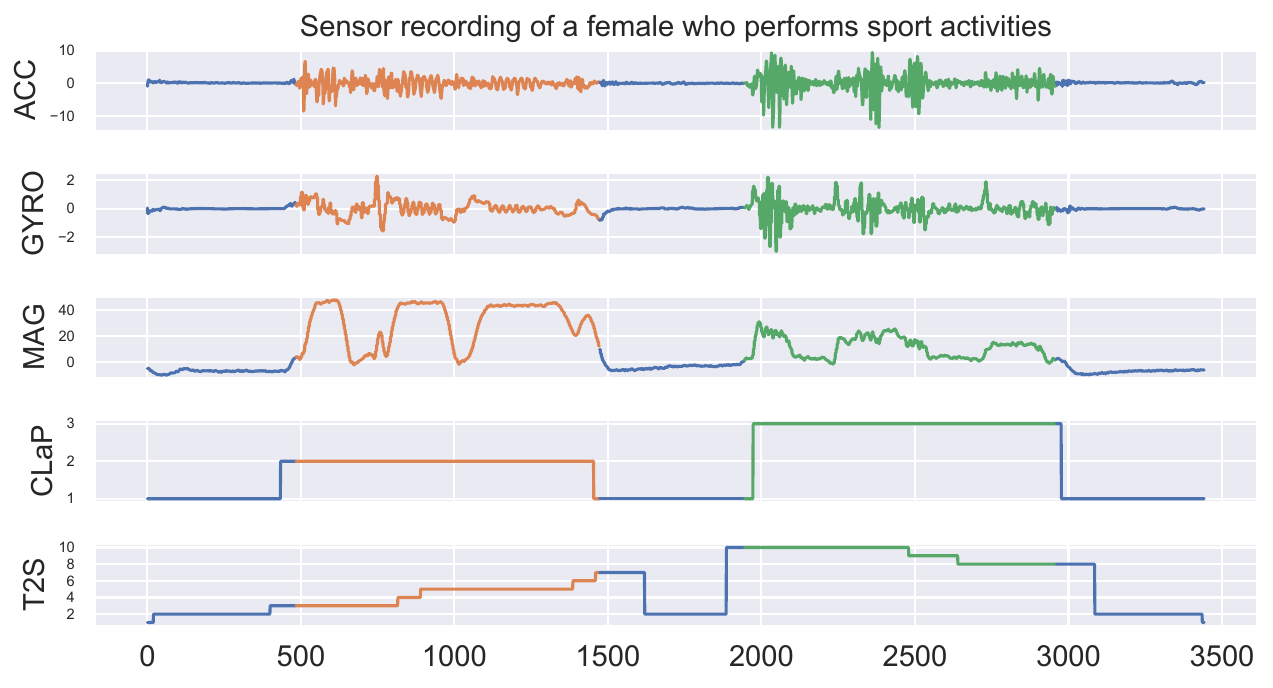}
    \caption{Top-3: human activity recording showing X-axis acceleration, gyroscope, and magnetometer measurements~\cite{ErmshausHumanAS} of different motions. Activities are consecutive, differently coloured sequences. 2nd from bottom: state sequence predicted by CLaP, assigning each data point one of three possible activity labels (1 for resting, 2 for squats, and 3 for loosening legs). Bottom: state sequence from competitor method Time2State, with 10 (too many) predicted labels.\label{fig:sport_activity_example}
    }
\end{figure}

A popular approach to TSSD is a combination of TS segmentation (TSS) and clustering techniques~\cite{Wang2024UnsupervisedTS}. A baseline approach could, for example, use the ClaSP (Classification Score Profile) algorithm~\cite{Ermshaus2022ClaSP} to compute a TSS and cluster the resulting segments using Time2Feat~\cite{Bonifati2023InterpretableCO} to predict state labels. The central limitation of this pipeline is its reliance on distance calculations in clustering, which are known to show limited performance compared to supervised methods used in TSC~\cite{Middlehurst2023BakeOR}. Unsupervised distance-based methods determine the state membership of segments solely within the feature space, whereas supervised techniques could also utilize the label space in this computation. This issue prevails to all state-of-the-art methods~\cite{Matsubara2014Autoplait,Johnson2010TheHD,Hallac2017ToeplitzIC,Wang2023Time2StateAU,Lai2024E2UsdEU}.

% Early state-of-the-art methods use Hidden Markov Models (HMMs) to model and label individual data points~\cite{Matsubara2014Autoplait,Johnson2010TheHD}. More recent approaches learn partial correlation structures and deep learning embeddings from subsequences~\cite{Hallac2017ToeplitzIC,Wang2023Time2StateAU}. However, current approaches are limited due to their reliance on common unsupervised methods, which reach their limits when tackling the complexity of TSSD.

% these methods require careful tuning of use-case-dependent hyper-parameters (e.g., Time2State~\cite{Wang2023Time2StateAU}), output inaccurate segmentations (e.g., HDP-HSMM~\cite{Johnson2010TheHD}, TICC~\cite{Hallac2017ToeplitzIC}), or cannot sufficiently model complex real-world TS (e.g., AutoPlait~\cite{Matsubara2014Autoplait}). 

We present CLaP (Classification Label Profile), a domain-agnostic TSSD algorithm that uses self-supervised change analysis to leverage the predictive power of supervised classification algorithms to the unsupervised TSSD problem. It is accurate, scalable, and capable of processing both univariate and multivariate TS. CLaP is hyper-parameter-free and learns all information from the data at hand. The routine first computes a segmentation with ClaSP, which internally uses a binary self-supervised classifier to differentiate segments based on differently shaped subsequences. CLaP then cross-validates a multi-class classifier with segment-labelled subsequences to measure the confusion between segments. This information is central to labelling segments that share the same state or not. In a subsequent merging procedure, the algorithm iteratively combines confused labels, representing the same state, using a novel merge criterion called classification gain. By design, CLaP automatically determines the number of states in a TS; an otherwise difficult to set hyper-parameter. 

% However, it allows for the incorporation of domain knowledge, if available, to guide its TSSD. 

Unlike competing approaches, which use classical unsupervised techniques, CLaP is the first method to tackle the TSSD problem with supervised learning methods, by explicitly framing it as a self-supervised task. Segments are modelled directly in label space, which is more discriminative than feature space, used in earlier algorithms. CLaP introduces a novel merging procedure that learns states by fusing segment labels according to their mutual confusion. This process is guided by a new score that enables fair comparisons across classification problems of differing complexity.

% It creates an interpretable annotation, the classification label profile, where each segment is represented by a flat line, at the level of its state. This constitutes the predicted state sequence.

Figure~\ref{fig:sport_activity_example} shows an example of how CLaP and a state-of-the-art competitor Time2State~\cite{Wang2023Time2StateAU} annotate a multivariate time series (MTS) with a state sequence. The recording captures smartphone sensor readings of a 30-year-old female resting (blue), performing squats (orange), resting again (blue), loosening her legs (green), and resting again (blue)~\cite{ErmshausHumanAS}. CLaP accurately identifies the transitions between activities and correctly assigns 3 distinct labels to the 5 segments (segments with the same activity are visualized as equally high flat lines; ones with different activities appear as lines at varying levels in the visualization). Time2State, in comparison, correctly identifies the resting periods, but overestimates the total number of activities being 10 separated by 12 transitions.

% It segments the raw data into states, enabling both individual analysis and inspection of transitions between activities.

In summary, this paper's contributions are:

\begin{enumerate}
    \item We present CLaP, a new domain-agnostic and hyper-parameter-free TSSD algorithm that leverages self-supervised TS classification for segmentation and labelling to predict state sequences of multivariate TS. We provide technical descriptions, a computational complexity analysis, examples, and a Python implementation.

    % By default, CLaP automatically learns suitable parameters from the data, but can incorporate domain knowledge (e.g., CPs, state labels) if available. 

    \item We propose two technical novelties that make TS classifiers applicable for TSSD: confused merging, a mechanism to reduce a TS state labelling to a minimal required set of labels; and classification gain, which quantifies the increase in F1 score of a classification compared to a random one.

    \item We assess the accuracy and runtime of CLaP and six state-of-the-art competitors (Time2State, HDP-HSMM, E2USD, ClaSP2Feat, TICC, AutoPlait) on $405$ TS from five benchmarks. CLaP significantly outperforms all rivals in accuracy and has the best accuracy-runtime tradeoff. It achieves the highest mutual information of $60.4$\% with ground truth annotations, an increase of $16.1$ percentage points (pp) compared to the second-best competitor Time2State.
\end{enumerate}

To foster the reproducibility of our findings, we created a supporting website~\cite{CLaPWebpage} that contains all source codes, our evaluation framework, Jupyter notebooks for exploratory analysis, raw measurements, scores, and visualizations. The remaining paper is organized as follows: Section~\ref{sec:background} introduces the necessary definitions and background of this work. In Section~\ref{sec:clap}, we present CLaP and its technical novelties in detail. Section~\ref{sec:evaluation} shares the results of extensive accuracy and runtime experiments of CLaP and the competitors. In Section~\ref{sec:related-work}, we review related works, and Section~\ref{sec:conclusion} concludes.

\section{Definitions and Background} \label{sec:background}
% processes, ts, subsequences, state detection, tsc

We formally define the concepts of abstract processes, time series, subsequences, state sequences, the time series state detection (TSSD) problem, and introduce ideas of TS change analysis.

\begin{definition}
An \emph{abstract process} $P = (S,R)$ consists of one or more discrete and distinct states $s_1, ..., s_l \in S$ that are pairwise separated by transitions $(s_i,s_j) \in R \subseteq  S \times S$, with $s_i \neq s_j$.
\end{definition}

Following the definition of Wang et al.~\cite{Wang2024UnsupervisedTS}, states refer to distinct phases of real-world processes observable through sensor measurements. We assume each state has a stationary, recognizable property that persists over time and makes it distinguishable from other states. Examples include human activities or machine conditions. This definition excludes processes with homonym states, that emit indistinguishable signals, and synonym states, which cannot be recognized by observation. It also excludes states that have drifting semantics over time.

Transitions are, by definition, changes between states and constitute the links among them in processes. We assume that a transition leads to a change in the observations of a process. Gradual changes, such as trends, can be modelled as separate states. We make no assumptions about transition causes, i.e., whether they occur randomly or systematically.

For analysis, we consider measurements emitted by one or multiple sensors observing outcomes or byproducts of a process. Human activity, for instance, can be tracked by inertial measurement units (IMUs) in smartphones~\cite{Lara2013ASO} and industrial machinery can be monitored with IoT devices~\cite{Wang2020Apache}, which results in temporal data.

\begin{definition}
A \emph{time series} (TS) $T$ is an ordered sequence of $n \times d \in \mathbb{N}^2$ dimensional vectors $T=(\vec{t}_{1},\dots,\vec{t}_{n}) \in \mathbb{R}^{n \times d}$ that simultaneously measures $d$ observable outputs of a process $P$.
\end{definition}

Each $t_i$ has $d$ dimensions, or channels, one for each sensor. Its values, also called data points or measurements, are equi-distant and ordered by time, e.g. 1 data vector is recorded every 10 milliseconds. 

\begin{definition}
Given a TS $T$, a \emph{subsequence} $T_{s,e}$ of $T$ with start offset $s$ and end offset $e$ is the $d$-dimensional slice of contiguous observations from $T$ at position $s$ to position $e$, i.e., $T_{s,e} = (\vec{t}_s,\dots,\vec{t}_e)$ with $1\leq s \leq e \leq n$. The length of $T_{s,e}$ is $|T_{s,e}| = e-s+1$.
\end{definition}

We use the terms \emph{subsequence} and \emph{window} interchangeably, and refer to their length as the \emph{width}. A state from $S$ yields its observable properties as a subsequence with shapes or statistics distinguishable from others. We call these core structures \emph{temporal patterns}, since they provide the necessary information to recognize the same state and distinguish it from others. Temporal patterns can drift or suddenly change over time, indicating a switch from one process state to another. Note, however, that local parts of channels contribute differently to the signal, for instance in amplitude.

\begin{definition}
Given a TS $T$ of size $n$ that captures a process $P = (S,R)$, the corresponding state sequence $Q = (s_{t_1},\dots,s_{t_n}) \in S^n$ contains the states $s_{t_i} \in S$ that are captured at time points $t_i \in T$.
\end{definition}

A \emph{change point} (CP) denotes an offset $i \in [1,\dots,n]$ for $t_i \in T$ that corresponds to a transition between states $s_{t_{i-1}}$ to $s_{t_i}$ in $Q$, where $(s_{t_{i-1}},s_{t_i}) \in R$ and $s_{t_{i-1}} \neq s_{t_i}$. For notational convenience, we consider the first and last values in $T$ as CPs. We call the subsequence between two CPs a \emph{segment} with variable size. A \emph{segmentation} of $T$ is the ordered sequence of CPs in $T$, i.e., $t_{i_{1}},\dots,t_{i_{n}}$ with ${1\leq i_1<\dots<i_n\leq n}$ at which the process $P$ changes state. 

\begin{definition}
The problem of \emph{time series state detection} (TSSD) is to recover the latent state sequence $Q$ of a process $P$, only by analysing the time series $T$, emitted by $P$.
\end{definition}

The TSSD problem, as defined here, is unsupervised and twofold: We need to find a segmentation of $T$ that captures the state transitions $R$, as well as the distinct states $S$ from all segments, in order to predict a state sequence $\hat{Q}$. An optimal result yields a $\hat{Q}$ that is isomorphic to $Q$, because an algorithm does not have access to the actual state labels. To evaluate the quality of a predicted state sequence, we can measure its alignment with the ground truth annotated by domain experts — e.g., by inspecting Covering~\cite{van2020evaluation} or the mutual information~\cite{Nguyen2010InformationTM} of segments.

% An example is an activity routine, in which a human subject first waits, then walks, waits again, and goes on to perform other motions, as studied in~\cite{Reiss2011TowardsGA}. The TSSD task would be to annotate e.g. the triaxial acceleration signal of the subject's left calf, with a state sequence of zeros (for waiting), ones (for walking), zeros again (for waiting), and so forth; capturing the latent process.

\subsection{Self-supervised Change Analysis}

To label TS with their states, CLaP computes segments of $T$ and iteratively relabels them based on mutual similarity. Contrary to classical clustering approaches, we do not compare distances, but build on self-supervised TS change analysis, first proposed by Hido et al.~\cite{Hido2008UnsupervisedCA}, that we shortly introduce next. 

Self-supervised learning is an unsupervised learning variant, in which the data itself is used to generate supervision labels. Consider two data sets, $X_A$ and $X_B$, representing subsequences from different segments. We assign label $0$ to samples from $X_A$ and $1$ to ones from $X_B$, enabling a binary classification evaluation using cross-validation. A TS classifier, trained on the labelled subsequences, predicts labels for unlabelled instances. $k$-fold cross-validation evaluates the classifier by training it on $(k-1)$ parts of the data and testing it on the remaining one. This process repeats $k$ times, covering all combinations, with the average of the $k$ evaluation scores (e.g., F1-scores) representing the classifier's predictive power. This value measures the classifier's ability to distinguish between data sets $X_A$ and $X_B$. A high score implies high dissimilarity and unique characteristics between the segments, indicating they represent different states. Conversely, a lower score indicates similarities, suggesting the segments may belong to the same state.

\section{CLaP - Classification Label Profile} \label{sec:clap}

We propose the \emph{Classification Label Profile}, short \emph{CLaP}, a novel algorithm that formulates TSSD as a self-supervised classification problem. We first provide an overview and example of the main concept, before we explain in detail how to implement it for state detection in Subsections~\ref{sec:state-detection} to~\ref{sec:computational-complexity}.

\begin{definition}
Consider a process $P = (S, R)$, captured by a TS $T$ of size $|T| = n$, and a window size $w$. A CLaP is a tuple $(L, c)$ that annotates $T$ with a label sequence $L \in \{1, \dots, k\}^{n-w+1}$ and an associated cross-validation score $c \in [0,1]$.
\end{definition}

The label sequence $L$ links all the $n-w+1$ overlapping subsequences $T_{i,i+w-1}$ with one of $k$ labels, which are used to represent the states in $S$. CLaP is computed from a subsequence classification problem with $k$ classes, whose discriminatory power is summarised in the score $c$. We derive $L$ from a self-supervision mechanism (see Subsection~\ref{sec:state-detection}) and calculate $c$ as the F1-score from a 5-fold cross-validation with the ROCKET classifier~\cite{Dempster2019ROCKETEF}, using the subsequences from $T$ as data and $L$ as artificial ground truth labels. Low scores of $c$ indicate CLaP is not able to accurately differentiate similar from dissimilar subsequences, while high scores show that it constitutes distinguishable windows.

% The labelling in CLaP directly affects its performance. 

\begin{figure}[t]
    \includegraphics[width=1.0\columnwidth]{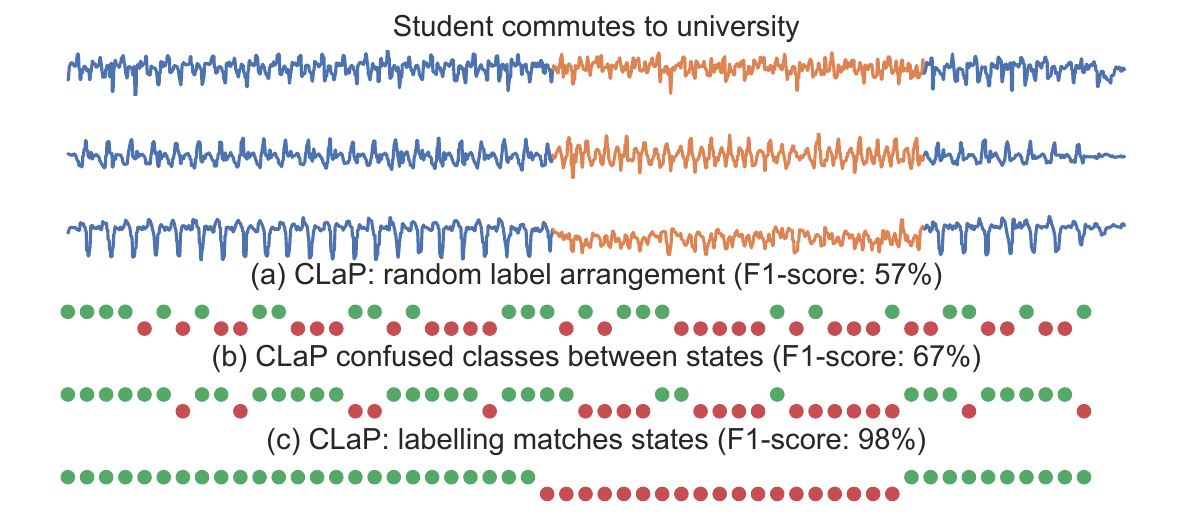}
    \caption{Top: excerpt of triaxial mobile phone acceleration recording from male student commuting to university, showing horizontal (blue) and downstairs (orange) walking motions~\cite{ErmshausHumanAS}. 4th from top to bottom: three CLaPs with different labellings (green and red) and associated F1 cross-validation scores. Profiles (a to c) incrementally align more with the true state sequence, which leads to increasing F1 scores.\label{fig:clap_example}
    }
\end{figure}

Figure~\ref{fig:clap_example} shows an example of three different CLaPs. The top part illustrates a triaxial human activity recording from a 23-year-old male, switching between horizontal (blue) and downstairs (orange) walking. The true state sequence of this recording assigns, for example, 1s to the blue data points and 2s to the orange measurements. The bottom part displays three different CLaPs (a to c). The sequence of green and red dots shows the label configuration $L$, with the associated F1-score $c$ annotated. CLaP (a) presents a random label configuration that does not match the true state sequence, resulting in poor cross-validation performance (57\% F1-score). The profile in (b) somewhat aligns with \(Q\) but confuses some instances (67\% F1-score). In contrast, CLaP in (c) perfectly resembles the true state sequence of the human activity recording, leading to a near-optimal performance (98\% F1-score).

The true state sequence $Q$ of a TS $T$ constitutes a CLaP that scores the highest discriminatory power compared to all possible sequence permutations. Our implicit assumptions thereby are three-fold: (1) Process states can be conceptualized as classes of TS windows, (2) subsequences (of appropriate length) can serve as class exemplars with distinctive features, (3) and a TS classifier cross-validation score correlates with the quality of the label configuration. With these concepts, we can tackle an unsupervised problem with the predictive power of TS classification (TSC) algorithms. Much research interest continues to increase accuracy and efficacy for the task (see Middlehurst et al.~\cite{Middlehurst2023BakeOR} for a survey). However, searching a high-quality state sequence prediction $\hat{Q}$ using CLaP is non-trivial.

The naive approach of enumerating all $k^{n-w+1}$ candidates and choosing the highest-scoring one (according to $c$) is computationally inefficient and yields further challenges. For instance, the number of classes $k$ is typically not known in advance. Also, the large collection of label combinations makes the cross-validation score incomparable for different values of $k$. We propose efficient and accurate solutions to prune the candidates and make their scores comparable in Subsections~\ref{sec:state-detection} and~\ref{sec:classification-gain}, bridging the gap to enable TSC methods to be directly applicable for TSSD. Subsection~\ref{sec:computational-complexity} analyses the runtime and space complexity of CLaP.

% The ''best'' CLaP may not even capture states, but rather recurring patterns (e.g., motifs) distributed throughout $T$

\begin{algorithm}[t]
	\caption{Classification Label Profile}\label{alg:clap}
	\begin{algorithmic}[1]
		\Procedure{clap}{$T$}		  
			\State $w \gets \textsc{learn\_subsequence\_width}(T)$ \Comment{Run SuSS}
            \State $cps \gets \textsc{compute\_segmentation}(T)$ \Comment{Run ClaSP}
            \State $X, L \gets \textsc{create\_data\_set}(T, w, cps)$
			\State $y_{pred} \gets \textsc{cross\_val\_clf}(X, L)$ \Comment{5-fold ROCKET CV}

            \For{$i \in [1, \dots, \|cps\|]$}
                \State $merged \gets false$
                \State $labels, conf \gets \textsc{calc\_confused\_labels}(L, y_{pred})$

                \For{$(l_1, l_2) \in \textsc{rank}(labels, conf)$}
                    \State $\hat{L} \gets \textsc{replace}(L, l_1, l_2)$ \Comment{Merge labels}
                    \State $\hat{y}_{pred} \gets \textsc{replace}(y_{pred}, l_1, l_2)$
                
                    \If{$\textsc{cgain}(\hat{L}, \hat{y}_{pred}) \geq \textsc{cgain}(L, y_{pred})$}
                        \State $L, y_{pred} \gets \hat{L}, \hat{y}_{pred}$ \Comment{Update labels}
                        \State $merged \gets true$
                        \State \textbf{break}
                    \EndIf
                \EndFor

                \State \textbf{if} $merged \neq true$ \textbf{then break} \Comment{Early stopping}
            \EndFor
   
			\State \Return{$(L, \textsc{score}(L, y_{pred}))$} \Comment{Calc. F1-score}
		\EndProcedure
	\end{algorithmic}
\end{algorithm}

\subsection{Confused Merging} \label{sec:state-detection}

\begin{figure}[t]
    \includegraphics[width=1.0\columnwidth]{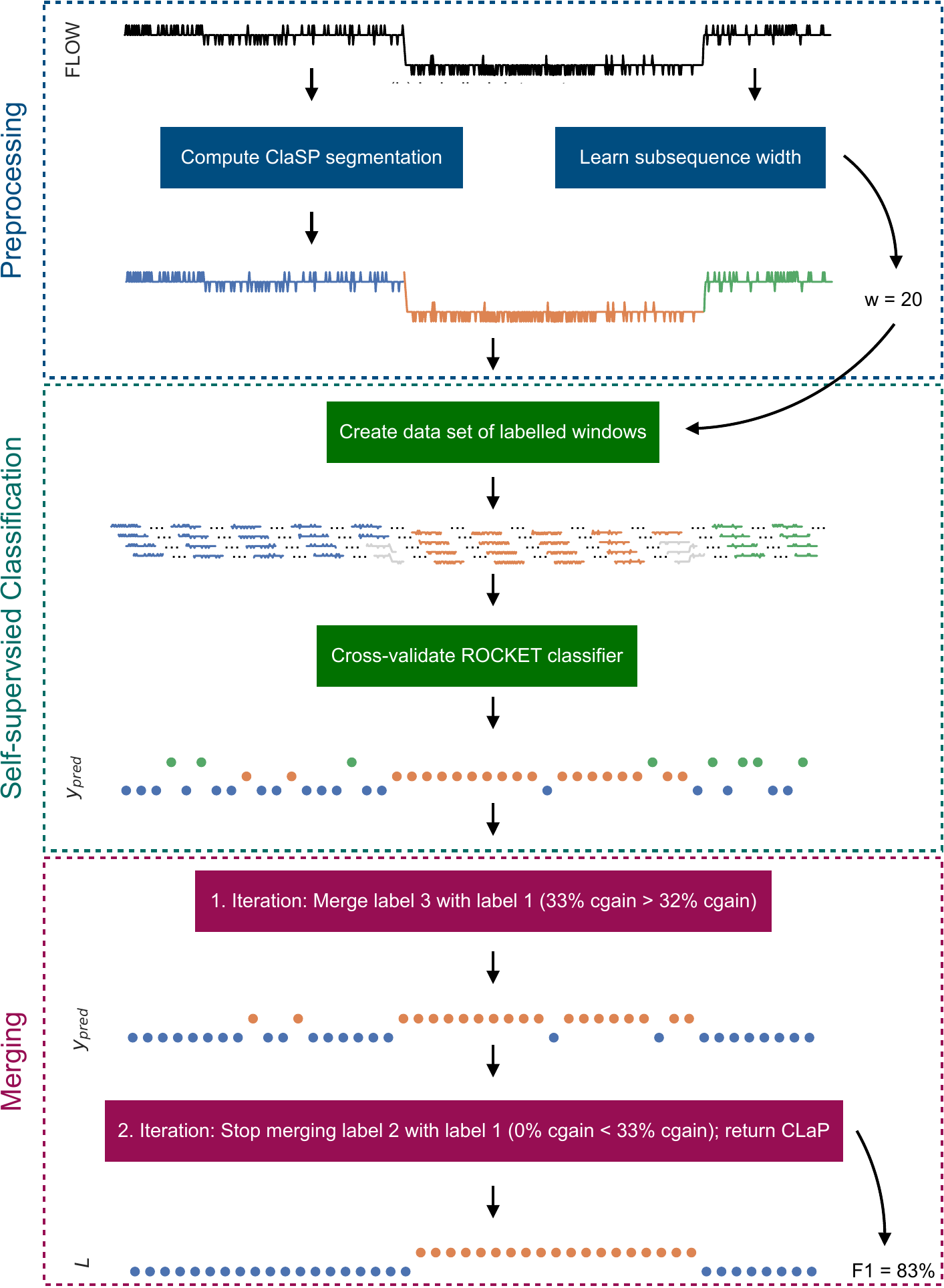}
    \caption{Workflow of CLaP for a water circuit~\cite{skab}. TS captures fluid circulation flow with opened (blue and green) and closed valve (orange). Preprocessing: ClaSP segmentation divides TS into three segments, SuSS learns the subsequence width. Self-supervised Classification: labelled subsequences constitute the data set; grey ones are disposed; ROCKET classifier computes initial predicted cross-validation labels. Merging: labels are combined based on confusion as long as cgain increases. Final CLaP with artificial ground truth labels and F1-score of 83\% is returned. \label{fig:clap_workflow}
    }
\end{figure}

The TSSD problem can be divided into a segmentation followed by clustering. We use this formulation to drastically prune the amount of potential CLaPs. First, we compute a segmentation of a TS $T$. This enables us to label only few segments, instead of many single data points. Secondly, we propose a novel agglomerative self-supervision mechanism, that greedily clusters confused state labels with a new bespoke merge criterion. The entire process, called \emph{Confused Merging}, combines several design choices, that we fixed using ablations (see Subsection~\ref{sec:ablation-study}). For each component, we considered multiple domain-agnostic methods and chose the best-performing one. CLaP is hyper-parameter-free, automatically learns the number of $k$ classes and stops as soon as no classes must be merged. Pseudocode is provided in Algorithm~\ref{alg:clap} and the workflow is illustrated in Figure~\ref{fig:clap_workflow}.

\textbf{Preprocessing:} Algorithm~\ref{alg:clap} receives a TS $T$ as user input (line~1) and starts by learning its subsequence width $w$ (line~2), needed to compute the sliding window in CLaP. Multiple techniques are available for this task and based on the idea that temporal patterns of similar size repeat throughout TS, an assumption we share in CLaP. We choose SuSS (Subsequence Summary Statistics), one of the best-performing procedures, according to~\cite{Ermshaus2022WSS}. For multidimensional TS, we compute one window size per dimension and then compute their average. The CLaP algorithm continues to compute a segmentation of $T$ with the ClaSP method~\cite{Ermshaus2022ClaSP} (line~3), which can partition multivariate TS~\cite{Ermshaus2024MultivariateClaSP} from different domains~\cite{Ermshaus2023TimeSS,Ye2024DiffusionbasedMF} and has an extension for very large and streaming data~\cite{Ermshaus2024ClaSS}. This preprocessing step removes the need to label individual data points in CLaP and instead only requires assigning classes to entire segments. 

\textbf{Self-supervised Classification:} The procedure creates an initial labelled data set from $T$ by creating a sliding window $X$ (width of $w$, stride of $\frac{w}{2}$) and labelling the resulting subsequences with their corresponding segment ranks in $L$, i.e. label 1 for the first segment, label 2 for the second one, label 3 for the third, and so on (line~4). Subsequences that overlap neighbouring segments with at least $\frac{w}{2}$ values are disposed. The motivation behind this labelling is twofold: it encodes the CPs (from the segmentation) as a labelling and first assumes all segments to capture distinct states. This enables the later merging, which fuses labels from segments sharing state. CLaP evaluates the ROCKET classifier, which is fast and accurate~\cite{Bagnall2020Usage}, with the labelled data set in a 5-fold cross-validation and stores the predicted labels in $y_{pred}$ (line~5). This creates an initial measure for the quality of the labelling. Compared to the artificial ground truth $L$, segments representing single states are expected to mostly match. Conversely, multiple segments sharing state, are expected to confuse labels, facilitating their identification. 

\textbf{Merging:} CLaP uses the initial labelled data set $(X, L)$ and its cross-validated prediction labels $y_{pred}$ to iteratively fuse segments representing the same state (lines~6--19). To achieve this, the procedure greedily merges highly confused labels, as their associated subsequences are difficult to separate. For every unique label $l_1$ from $L$, CLaP determines its most confused counterpart $l_2$, according to $y_{pred}$ (line~8). This information is directly derived from a confusion matrix. The algorithm iterates the most confused label pairs with descending confusion to check if they can be merged (lines~9-17). Specifically, it creates temporary label vectors $\hat{L}$ and $\hat{y}_{pred}$, where $l_2$ is replaced with $l_1$ (lines~10--11), and checks if their classification gain (see Subsection~\ref{sec:classification-gain} for details) is greater (or equal) to the one from $L$ and $y_{pred}$ (line~12). In this case, the method updates the current with the merged labelling (line~13) and stops merging; otherwise, it continues with less confused label pairs. After a successful merge iteration, CLaP starts another round with updated artificial ground truth and predicted labels. Once no label pair can be further merged (line~18), the procedure stops and returns the current label configuration $L$ and its associated cross-validation score, according to $y_{pred}$, constituting CLaP (line~20). The central mechanism of this process is confused label pair selection. We exploit the fact that segments sharing state are confused by the classifier, because of similar temporal patterns that are differently labelled. Hence, we use it as a selection criterion, implementing it in an agglomerative fashion, to learn the amount of states and stop as soon as possible. Classification gain (see Subsection~\ref{sec:classification-gain}) regularises the merging to ensure that only labels from sufficiently similar segments are fused.

\textbf{Workflow:} Figure~\ref{fig:clap_workflow} illustrates the confused merging algorithm for water flow circulation in a testbed~\cite{skab}. The preprocessing (blue frame) divides the TS into three segments (blue, orange, green), capturing two states; namely open and closed valve. It also computes the subsequence width of 20 with SuSS. The self-supervision mechanism (green frame) creates labelled subsequences using segment ranks (1, 2, 3) and uses ROCKET to compute initial predicted cross-validation labels $y_{pred}$ that show high confusion between label 1 and 3. Merging (purple frame) finds label 3 to have the highest overall confusion (with label 1) and fuses both, increasing classification gain (from 32\% to 33\%). Merging even further to one class would reduce cgain to 0\%, hence the procedure stops. The resulting artificial ground truth $L$ and it's 83\% F1-score constitute CLaP and correctly capture the latent states.

The procedure is hyper-parameter-free and it learns two model-parameters from the input TS: the subsequence width $w$ and the number of classes $k$. By default, CLaP uses five folds for cross-validation and draws at most 1k randomly sampled, labelled subsequences from the full data set to control runtime.

\subsection{Classification Gain} \label{sec:classification-gain}

% idea: plot that shows difference between random and actual classifciatiton for different number of classes

Confused merging relies on a merge criterion that decides if two confused classes should be combined or not. In agglomerative clustering, this is typically achieved by thresholding the distance computation. For CLaP, we could translate that into enforcing the F1-score (or e.g. cross-entropy loss) of the label configurations (Algorithm~\ref{alg:clap}, line~12) to pass a minimal (maximal) value. However, this is disadvantageous for three key reasons: (a) the F1-score (also accuracy, ROC/AUC, or entropy) depends on the classifier and data; rendering threshold selection use case dependent. (b) Comparing F1-scores of classification problems with different label distributions is meaningless; more labels increase problem complexity, resulting in generally lower F1-scores. (c) The F1-score does not reflect the improvement of one labelling over another, just their total qualities. To address these challenges, we propose a new merge criterion called \emph{Classification Gain}, which measures the increase in F1-score over its expected random score, thereby normalizing for problem complexity. It isolates predictive power from classification difficulty, facilitating performance comparisons across different problems.
\begin{definition}
    Given a ground truth labelling $y_{true} \in U^n$ of size $n$, with unique classes $U \subset \mathbb{N}$, the random F1-score is defined as: 
    \begin{align}
        f1_{rand}(y_{true}) &:= \frac{1}{\|U\|} \sum_{l \in U} \frac{2 \cdot TP(l)}{2 \cdot TP(l) + FN(l) + FP(l)}
    \end{align}
\end{definition}
The random (macro) F1-score provides an unbiased estimate of classification difficulty based on the label distribution $y_{true}$. It can be computed efficiently using probability theory in the following way: in a random classification, we expect each instance labelled $l \in U$ to be mapped to its own (or any other) class according to its likelihood of occurrence.
\begin{align}
    P(y_{true} = l) &:= \frac{\#(y_{true} = l)}{\|y_{true}\|} \\
    P(y_{true} \neq l) &:= 1 - P(y_{true} = l) = \frac{\#(y_{true} \neq l)}{\|y_{true}\|}
\end{align}
These class priors enable us to define the true positives (TPs) as the expected amount of instances that are randomly assigned to their class (Equation~4), the false negatives (FNs) as the expected number of samples that are mapped to another class (Equation~5), and the false positives (FPs) as the expected amount of misclassified examples that actually belong to a given class (Equation~6). 
\begin{align}
    TP(l) &:= \#(y_{true} = l) \cdot P(y_{true} = l) \\
    FN(l) &:= \#(y_{true} = l) \cdot P(y_{true} \neq l) \\
    FP(l) &:= \#(y_{true} \neq l) \cdot P(y_{true} = l)
\end{align}
Given the random macro F1-score for $y_{true}$ as an estimate of problem difficulty, we can decouple it from the performance of a classification to make it comparable across problems.
\begin{definition}
    Given a ground truth labelling $y_{true} \in U^n$ and associated predictions from a classifier $y_{pred} \in U^n$ of size $n$, the classification gain, short \emph{cgain}, is defined as: 
    \begin{align}
        cgain(y_{true}, y_{pred}) := f1(y_{true}, y_{pred}) - f1_{rand}(y_{true})
    \end{align}
\end{definition}
It reflects the improvement in macro F1-score of the classification over the expected random one. The central idea of $cgain$ is to normalize F1-score with the associated classification problem complexity to make the measure comparable for different label distributions. By comparing $cgain$ in CLaP (line~12), we enforce that the merge operation steadily improves classification quality, while not relying on thresholds. This addresses the problems (a) to (c) that suffer from greedily maximising performance.

Note that cgain is different from gini gain~\cite{Breiman2004TechnicalNS}, which measures the discrepancy in purity between data sets. Classification gain measures the quality for a given classification considering its difficulty, estimated by random classification.

\begin{figure}[t]
    \begin{minipage}{8cm}
        \includegraphics[width=1.0\columnwidth]{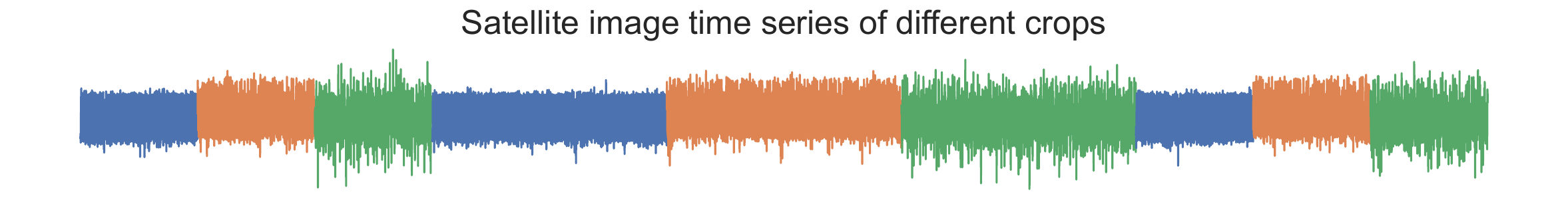}
    \end{minipage}
     \begin{minipage}{4cm}
        \includegraphics[width=1.0\columnwidth]{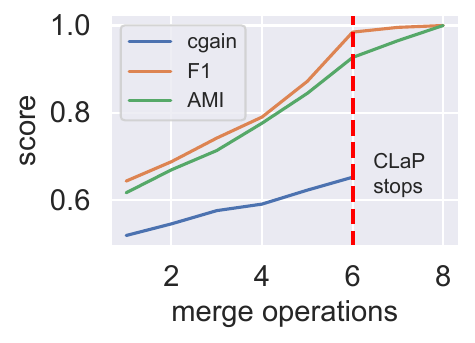}
    \end{minipage}
     \begin{minipage}{4cm}
        \includegraphics[width=1.0\columnwidth]{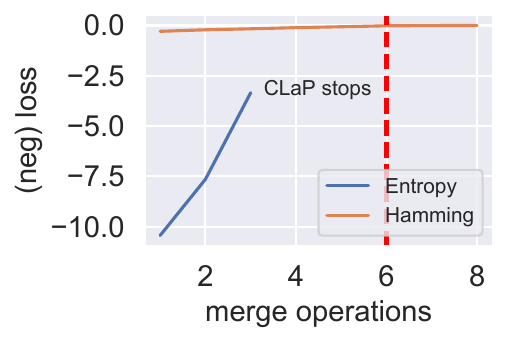}
    \end{minipage}
    \caption{Top: TS capturing 9 time spans of 3 different crops (blue, orange, green)~\cite{Schfer2021ClaSPT}. Bottom: CLaP run on TS with different merge criteria (scores and negative losses); dashed red line marks optimal amount of 6 merge operations, reducing 9 initial classes to 3. Only cgain accomplishes this; other measures misjudge label configuration quality and either merge too few (Entropy) or too many (F1, AMI, Hamming) classes.\label{fig:cgain}
    }
\end{figure}

\textbf{Example:} Figure~\ref{fig:cgain} (top) illustrates a satellite image TS capturing sensor observations of 3 different crops (coloured in blue, orange, green) as 9 segments. A correct segment state sequence (one label per segment) would be: 1,2,3,1,2,3,1,2,3. CLaP tries to find such sequence by merging classes from its initial sequence: 1,2,3,4,5,6,7,8,9. In Figure~\ref{fig:cgain} (bottom) we use different merge criteria in CLaP (line~12) for this computation and show their values for the merge operations. Generally, merging should increase the measures for 6 iterations (red dashed line) and then stop, as they begin to decrease. It protrudes that F1-score, adjusted mutual information score (AMI) and (negative) Hamming loss steadily increase, reducing the state sequence with 8 merges to just one state, namely 1. This is due to the fact that none of the measures considers problem complexity and overestimates label configuration quality with decreasing number of classes. Negative cross-entropy loss underestimates quality and merges only 3 times, as its value would decrease for a fourth iteration (line~12), resulting in the sequence: 1,2,3,1,2,4,1,5,6. Our proposed classification gain is the only criterion that merges exactly 6 times and would decrease in value thereafter; producing the ground truth result and stopping automatically.

%\subsection{Incorporating Domain Knowledge} \label{sec:domain-knwoledge}

%CLaP is, by design, hyper-parameter-free and unsupervised, making it advantageous for fast exploratory data analysis as well as being a component of automated data engineering workflows. However, experts often have domain knowledge, which could increase the quality of state detection. For example, a user may be able to indicate some CPs or state labels of single measurements in TS from its visual inspection or access to exogenous data. To make this knowledge accessible to CLaP, we allow the user to provide it as optional parameters, that we integrate in the computation.

%The CLaP method can take a (possibly) incomplete list of CPs and performs its preprocessing as described. However, after CPs are determined by ClaSP (Algorithm~\ref{alg:clap}, line~3), we add the user's CPs that are at least $5 \cdot w$ data points distant from other CPs; inhibiting duplicates and successive splits. Similarly, the procedure can receive a second (possibly) incomplete list of (offset, label) tuples, assigning single time points of the TS to state labels. After the labelled data set is created (line~4), we merge unique labels that the user denotes to come from single states. Thereafter, CLaP resumes as specified and performs confused merging.

%The provision of domain knowledge through optional parameters makes the central parts of CLaP configurable for experts and enables an active human-in-the-loop setting for the continued refinement of its results over multiple runs.

\subsection{Computational Complexity} \label{sec:computational-complexity}

The scalability of state detection algorithms is critical as the size of sensor recordings steadily grows. The complexity of CLaP mainly depends on the segmentation (Algorithm~\ref{alg:clap}, line~3), self-supervised classification (line~5) and merging (lines~6-19).

For a TS $T$ with $d$ channels of size $n$, the initial computation of the window size $w$ (line~2) with SuSS is in $\mathcal{O}(d \cdot n \log w)$~\cite{Ermshaus2022ClaSP}. Locating $c$ CPs in $T$ with ClaSP (line~3) is in $\mathcal{O}(d \cdot n^2 \cdot c)$, which involves cross-validating a self-supervised $k$-nearest neighbour classifier multiple times. The entire preprocessing complexity is in $\mathcal{O}(d \cdot n \log w + d \cdot n^2 \cdot c)$ which reduces to $\mathcal{O}(d \cdot n^2 \cdot c)$ as $w \leq n$.

Creating the data set for self-supervised classification (line~4) is in $\mathcal{O}(d \cdot n \cdot w)$, as it only requires reorganising $T$. Running a single validation of the ROCKET classifier is also in $\mathcal{O}(d \cdot n \cdot w)$; it convolves the data set with a fixed set of $10k$ kernels for feature generation and uses ridge regression for classification~\cite{Dempster2019ROCKETEF}. The same complexity holds for repeating the procedure 5 times during cross-validation (line~5), which leads to a total runtime in $\mathcal{O}(d \cdot n \cdot w)$ for the entire step. Note, that ROCKET's runtime hides five-digit constants for calculating convolutions with $10k$ kernels~\cite{Dempster2019ROCKETEF}.

The confused merging calculates a confusion matrix (line~8) for the $\lfloor \frac{2 \cdot (n - w)}{w} \rfloor + 1$ labels from the cross-validation in $\mathcal{O}(n)$; which only involves incrementing counts. The same complexity holds for replacing labels (lines~10--11) and calculating classification gain (line~12), which is mainly based on computing the confusion matrix for the random F1-score; counting labels and calculating likelihoods. This process is repeated, in the worst case, for $c+1$ confused label pairs in descending order which requires $\mathcal{O}(n \cdot c \log c)$ time. The entire process (lines~6--19) can be repeated up to $c$ times, resulting in a total runtime complexity of $\mathcal{O}(n \cdot c^2 \log c)$.

Considering the sum of all steps, CLaP requires $\mathcal{O}(d \cdot n^2 \cdot c) + \mathcal{O}(d \cdot n \cdot w) + \mathcal{O}(n \cdot c^2 \log c)$ time, which simplifies to $\mathcal{O}(d \cdot n^2 \cdot c + n \cdot c^2 \log c)$ for computing $n$ state labels with maximal $c+1$ unique classes. This runtime complexity allows for CLaP's application to medium-sized and large TS in the batch setting. For streaming purposes, however, further advancements are needed. 

The space complexity mainly depends on the data set size and confusion matrix, which is in $\mathcal{O}(d \cdot n \cdot w + c^2)$.

\section{Experimental Evaluation} \label{sec:evaluation}

We evaluated the accuracy, runtime, and scalability of CLaP against six competitors on five large benchmark data sets ($405$ TS) to assess its performance on real-world TS data. Subsection~\ref{sec:experiment-setup} presents the data sets, evaluation metrics, and competitors. In subsections~\ref{sec:comparative-analysis} and~\ref{sec:runtime}, we present accuracy and runtime results of CLaP and its competitors. Subsections~\ref{sec:ablation-study} and~\ref{sec:sensitivity} analyse the results of an ablation study with different design choices for CLaP and sensitivity experiments, with varying data quality. Finally, subsection~\ref{sec:discussion} discusses strengths and weaknesses of CLaP and subsection~\ref{sec:usecases} showcases a real-world use case. All experiments were conducted on an Intel Xeon 8358 with 2.60 GHz, 2 TB RAM, 128 cores, running Python 3.8. All of our experimental results, figures, raw measurements per data set, codes, Jupyter Notebooks, and TS used in the evaluation are available on our supporting website~\cite{CLaPWebpage} for inspection and to ensure reproducibility.

\subsection{Experiment Setup} \label{sec:experiment-setup}

\begin{table}[t]
    \begin{centering}
        \caption{Technical specifications of TS used in experiments.\label{tab:db_spec}}	
        \begin{tabular}{c|ccc}
            \toprule 			
            Name & No. & TS Length & No. States (Segs.)  \tabularnewline 
            & TS / Dims. & Min/Med./Max & Min/Med./Max \tabularnewline \hline
            TSSB~\cite{Ermshaus2022ClaSP} & 75 / 1 & 240 / 3.5k / 20.7k & 1 (1) / 3 (3) / 7 (9) \tabularnewline 
            UTSA~\cite{Gharghabi2018DomainAO} & 32 / 1 & 2k / 12k / 40k & 2 (2) / 2 (2) / 3 (3) \tabularnewline \hline
            HAS~\cite{ErmshausHumanAS} & 250 / 9 & 340 / 5k / 41.5k & 1 (1) / 3 (4) / 12 (15) \tabularnewline
            SKAB~\cite{skab} & 34 / 8 & 745 / 1.1k / 1.3k & 2 (2) / 2 (3) / 2 (3) \tabularnewline
            MIT-BIH~\cite{Moody2001TheIO} & 14 / 1 & 57k / 258k / 650k & 2 (3) / 2 (8) / 5 (16) \tabularnewline \hline
            Total & 405 / 9 & 240 / 4.8k / 650k & 1 (1) / 3 (3) / 12 (16) \tabularnewline
            \bottomrule 			
        \end{tabular}
    \end{centering}    
\end{table}

\textbf{Data Sets:} We evaluated CLaP and its competitors on a total of $405$ small to very large-sized (240 to 650k) TS from five public segmentation benchmarks (see Table~\ref{tab:db_spec}). Ground truth CPs and states, annotated by domain experts, are available and were used for evaluation. The benchmarks can be divided into two groups: cross-domain and single-domain use cases.

UTSA~\cite{Gharghabi2018DomainAO} and TSSB~\cite{Ermshaus2022ClaSP} consist of 75 and 32 univariate TS, representing a diverse collection of cross-domain problem settings featuring biological, mechanical, and synthetic processes from sensor, device, image, spectrogram, and simulation signals. Conversely, HAS~\cite{ErmshausHumanAS} contains 250 9-dimensional TS from smartphone sensors, capturing indoor and outdoor motion sequences from students. SKAB~\cite{skab} consists of 34 8-dimensional TS from machine sensors observing a water circulation testbed with pumps, tanks, and valves, undergoing both normal and anomalous conditions. MIT-BIH~\cite{Moody2001TheIO} features 14 univariate high-resolution ECG recordings from patients with normal cardiac activity and prolonged types of arrhythmias.

All TS combined capture versatile processes from natural phenomena, humans, animals, and industrial machinery. The number of distinct states per process varies between 1 and 12, occurring in 1 to 16 segments. In 65\% of all data sets, the number of states and segments match; in 15\% a single state reoccurs, 9\% show two reoccurrences, and the remaining 11\% exhibit three to 14 reoccurrences. It is worth noting that the data sets vary greatly in problem complexity, which tends to increase with the number of states and transitions in the processes.

\begin{table}[t]
    \begin{centering}
        \caption{List of competitors for experimental evaluation.\label{tab:competitors}}	
        \begin{tabular}{c|cc}
            \toprule 			
            Competitor & State Detection Method  \tabularnewline \hline
            AutoPlait~\cite{Matsubara2014Autoplait}  & Hidden Markov Models  \tabularnewline
            CLaP  & Self-supervision \tabularnewline
            ClaSP2Feat & Self-supervision \tabularnewline
            E2USD~\cite{Lai2024E2UsdEU} & Deep Learning \tabularnewline
            HDP-HSMM~\cite{Johnson2010TheHD}  & Hidden Markov Models \tabularnewline
            TICC~\cite{Hallac2017ToeplitzIC}  & Expectation Maximization \tabularnewline
            Time2State~\cite{Wang2023Time2StateAU}  & Deep Learning \tabularnewline
            \bottomrule 			
        \end{tabular}
    \end{centering}    
\end{table}

\textbf{Evaluation Metrics:} TSSD combines the segmentation and state identification tasks. We assessed both aspects separately with own scores, namely Covering~\cite{van2020evaluation} for segmentation and Adjusted Mutual Information (AMI)~\cite{Nguyen2010InformationTM} for state identification.

Covering measures how well the predicted and annotated segments overlap in location, disregarding the specific state labels of the segments. This allows for the comparison of segmentations of different sizes (including empty segmentations). Let the interval of successive CPs $[t_{c_i},\dots,t_{c_{i+1}}]$ be a segment in $T$, and let $segs_{pred}$ and $segs_{T}$ be the sets of predicted and annotated segmentations, respectively. Additionally, we set $t_{c_0} = 0$ as the first and $t_{c_{n+1}} = n+1$ as the last CP to include the first and last segments. The Covering score reports the highest-scoring weighted overlap between predicted and annotated segmentations (using the Jaccard index) as a normalized measure in the interval $[0,\dots,1]$, with higher scores being better (Equation~\ref{eqn:covering}).
\begin{align}
    \textsc{Covering} = \frac{1}{\|T\|} \sum_{s \in segs_{T}} \|s\| \cdot \max_{s' \in segs_{pred}} \frac{\| s \cap s' \|}{\| s \cup s' \|}
    \label{eqn:covering}
\end{align}

Mutual Information (MI) reports the similarity between two different labellings of the same data. It is invariant to specific label values and allows us to quantify the agreement between predicted and annotated states, disregarding their locations. The measure is adjusted to account for chance, as the score naturally increases with a larger number of states. Let $states_{pred}$ and $states_{T}$ represent the sets of predicted and annotated states, respectively, and let $H$ denote the entropy function. AMI measures the adjusted, weighted overlap between pairwise clusters as a normalized score within $[-1,\dots,1]$, with higher values indicating better agreement (Equation~\ref{eqn:ami}).

\begin{align} 
MI &= \sum_{s \in states_{T}} \sum_{s' \in states_{pred}} \frac{\|s \cap s'\|}{\|T\|} \log ( \frac{\|T\| \cdot \|s \cap s'\|}{\|s\| \cdot \|s'\|} ) \\ 
AMI &= \frac{MI - \boldsymbol{E}[MI]}{\textsc{avg}(H(states_{T}),H(states_{pred})) - \boldsymbol{E}[MI]}\label{eqn:ami} 
\end{align}

\begin{figure}[t]
    \begin{minipage}{8cm}
        \centering
        \includegraphics[width=0.9\columnwidth]{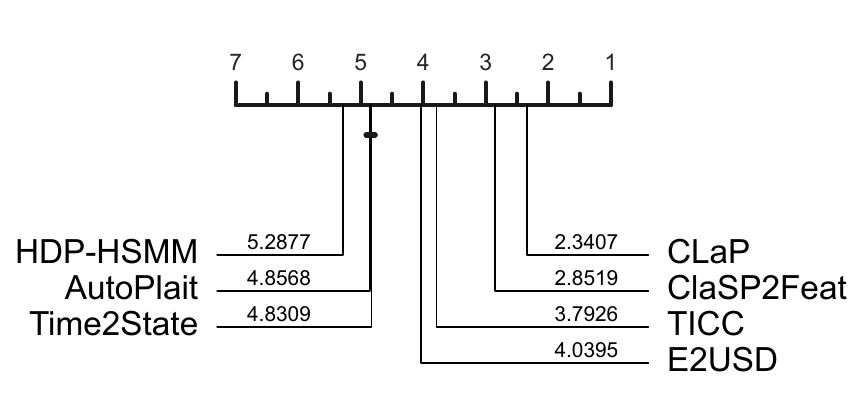}
    \end{minipage}
    \begin{minipage}{8cm}
        \centering
        \includegraphics[width=0.8\columnwidth]{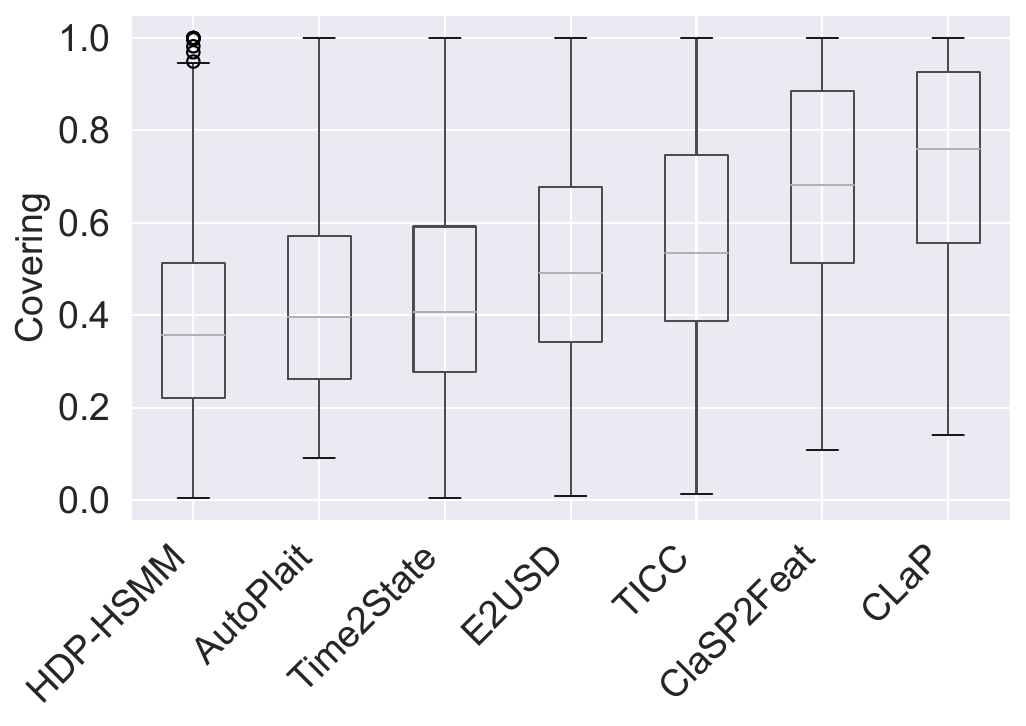}
    \end{minipage}
    \caption{Covering ranks (top) and box plot (bottom) on $405$ TS for CLaP (lowest rank) and six competitors.\label{fig:cd_benchmark_covering}}
\end{figure}

To compare different methods across multiple data sets, we aggregate scores into a single ranking. First, we compute the rank of the score for every algorithm and TS. Then, we average the ranks per method across data sets to obtain overall mean ranks. We visualize these values with Critical Difference (CD) diagrams~\cite{demvsar2006statistical} (see, e.g., Figure~\ref{fig:cd_benchmark_covering} top) which include an assessment of statistical differences between methods. The best approaches, with the lowest average ranks, are shown to the right of the diagram. Groups of algorithms that are not significantly different in rank are connected by a bar, based on a pairwise one-sided Wilcoxon signed-rank test with $\alpha=0.05$ and Holm correction. Besides CD diagrams, we discuss summary statistics, box plots~\cite{Tukey1977Exploratory}, and specific examples.

\begin{figure}[t]
    \begin{minipage}{8cm}
        \centering
        \includegraphics[width=0.9\columnwidth]{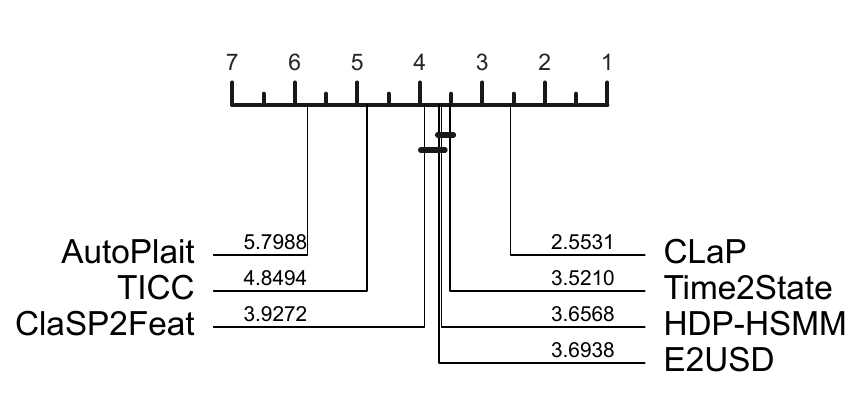}
    \end{minipage}
    \begin{minipage}{8cm}
        \centering
        \includegraphics[width=0.8\columnwidth]{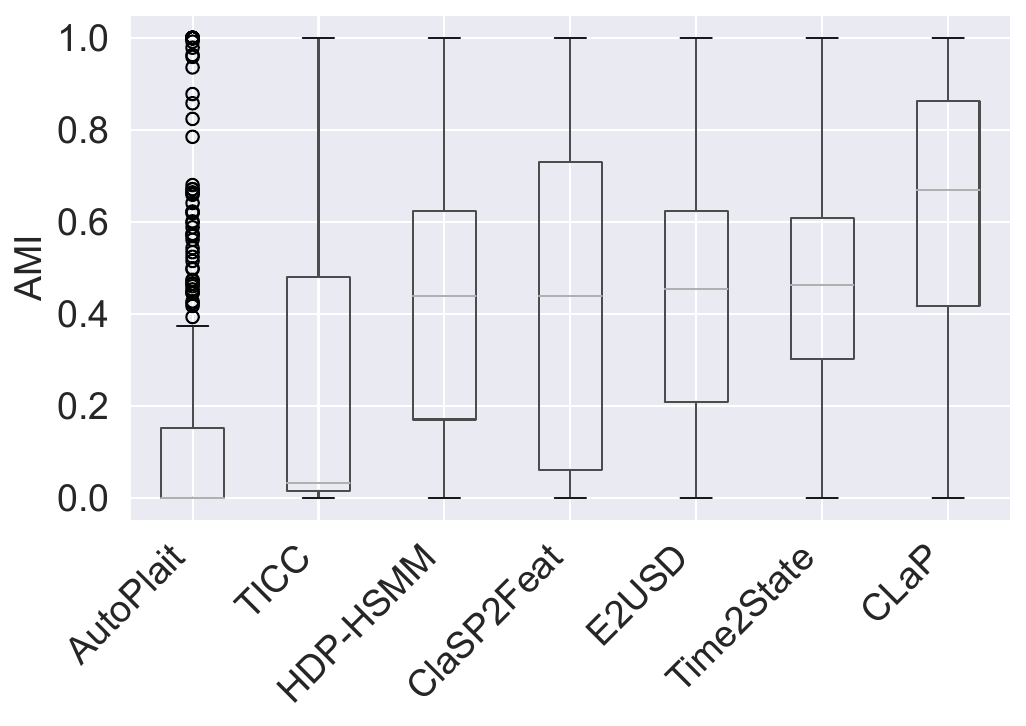}
    \end{minipage}
    \caption{AMI ranks (top) and box plot (bottom) for CLaP (lowest rank) and six competitors on $405$ TS.\label{fig:cd_benchmark_ami}}
\end{figure}

\textbf{Competitors:} We compare CLaP against six state-of-the-art state detection competitors (see Table~\ref{tab:competitors}). All implementations are openly available on our supporting website~\cite{CLaPWebpage}. As a baseline, we extract segments with ClaSP~\cite{Ermshaus2022ClaSP} and learn their labels using Time2Feat~\cite{Bonifati2023InterpretableCO}, which extracts and selects features from TS segments before clustering them. We call this approach ClaSP2Feat. We report results for two competitors based on Hidden Markov Models (HMMs). AutoPlait~\cite{Matsubara2014Autoplait} uses so-called multi-level chain models to learn states and applies the minimum description length (MDL) principle to automatically select the best-fitting model. HDP-HSMM~\cite{Johnson2010TheHD} is a Bayesian extension of traditional HMMs with efficient state sampling. Another statistical method we compare against is TICC~\cite{Hallac2017ToeplitzIC}, which formulates state detection as a covariance-regularized maximum likelihood problem and computes it using a greedy heuristic. From deep learning, we report results for Time2State~\cite{Wang2023Time2StateAU}, which learns an encoder from TS windows using a latent state encoding loss and clusters the embeddings with a Gaussian mixture model (DPGMM) to predict states. E2USD~\cite{Lai2024E2UsdEU} first creates a compact and informative TS embedding using Fast Fourier Transform (FFT) compression and decomposition techniques combined with contrastive learning, and then also uses DPGMM.

To prevent overfitting, we adopt the default hyper-parameters reported in the original publications or reference implementations: for HDP-HSMM we set $\alpha = 10k$ and $\beta = 20$; for TICC we use the annotated subsequence widths, $\lambda = 0.001$, $\beta = 2200$, a convergence threshold of $10^{-4}$, and a maximum of 10 iterations; Time2State and E2USD set a sliding window of 256 and step sizes of 10 and 50, respectively. HDP-HSMM, TICC, Time2State, and E2USD output only state sequences, so we derive CPs by locating offsets at changing states. ClaSP2Feat and TICC require the number of states as a hyper-parameter. We set this value from the ground truth annotations. Note that it provides an information advantage to both methods, which is, however, necessary to run them. All other competitors infer the number of states automatically.

% Besides the state detection competitors that can detect segment boundaries as well as state labels, we compare single aspects of CLaP against baselines from the respective subfields. For segmentation, we compare against BinSeg~\cite{Truong2020SelectiveRO}, PELT~\cite{Killick2011OptimalDO}, Window~\cite{Truong2020SelectiveRO} and FLUSS~\cite{Gharghabi2018DomainAO}, and RuLSIF~\cite{Hushchyn2020GeneralizationOC}. For clustering, we report results for $k$-Shape~\cite{Paparrizos2016kShapeEA}, $k$-Means, $k$-Medoids, agglomerative and spectral clustering~\cite{Holder2022ARA}.

\subsection{Comparative Analysis} \label{sec:comparative-analysis}

We evaluated the average (Covering / AMI) rank of CLaP and the six competitors on all $405$ TS to assess overall performance. Figures~\ref{fig:cd_benchmark_covering} and~\ref{fig:cd_benchmark_ami} (top) show the results. For both measures, CLaP (2.34 / 2.55) leads with a significant advantage, followed by ClaSP2Feat (2.85 / 3.93) and TICC (3.79 / 4.85) for Covering, and Time2State (4.83 / 3.52) as well as HDP-HSMM (5.29 / 3.66) for AMI. CLaP's lead in Covering, compared to ClaSP2Feat, confirms the substantial advantage of confused merging over clustering, as both methods utilize ClaSP for segmentation. Interestingly, the competitors' differences in average rank are not consistent across the measures. Considering the five benchmarks separately, CLaP also achieves first place for each collection, except in SKAB, where HDP-HSMM leads and CLaP scores third place because it detects wrongly located CPs that cannot be properly corrected by confused merging (see~\cite{CLaPWebpage}). CLaP's wins in Covering are not statistically significant, except for HAS. Conversely, its wins in AMI are statistically significant, except for MIT-BIH.

Looking at individual data sets, CLaP wins or ties $204$ / $193$ out of $405$ data sets (Covering / AMI), followed by ClaSP2Feat ($146$ / $94$) and TICC ($86$ / $62$). Note that the Covering counts exceed the total number of data sets because of ties (see~\cite{CLaPWebpage}). For the $37$ TS with only one state, TICC achieves the best results. However, for the $154$ data sets with two states and the $214$ TS with three or more states, CLaP leads by a significant margin. In a pairwise comparison of CLaP against every competitor, it outperforms each method in at least $38.3$\% / $56.0$\% of cases for Covering / AMI. In summary, the rankings show that CLaP on average surpasses state-of-the-art algorithms in terms of accuracy across the five benchmarks. 

\begin{table}[t]
    \begin{centering}
 	      \caption{Summary performances for CLaP (highest scores) and six competitors on $405$ TS.\label{tab:benchmark_summary}}
	    \captionof*{table}{Covering / AMI}
    		\begin{tabular}{c|ccc}
    			\toprule 			
    			 & average (in \%) & median (in \%) \tabularnewline
    			\hline 
              CLaP & $\textbf{72.9} \pm 21.6$ / $\textbf{60.4} \pm 30.9$ & $\textbf{76.0}$ / $\textbf{67.0}$ \tabularnewline
              ClaSP2Feat & $68.4 \pm 22.9$ / $44.1 \pm 34.4$ & $68.1$ / $43.9$ \tabularnewline
              TICC & $57.2 \pm 24.2$ / $24.4 \pm 34.8$ & $53.4$ / $3.1$ \tabularnewline
              E2USD & $50.0 \pm 23.3$ / $42.0 \pm 27.3$ & $49.1$ / $45.4$ \tabularnewline
              Time2State & $43.9 \pm 22.6$ / $44.3 \pm 23.3$ & $40.7$ / $46.3$ \tabularnewline
              AutoPlait & $45.8 \pm 25.0$ / $17.5 \pm 32.9$ & $39.6$ / $0.0$ \tabularnewline
              HDP-HSMM & $38.7 \pm 24.1$ / $41.5 \pm 27.9$ & $35.7$ / $43.9$ \tabularnewline
            \bottomrule 			
        \end{tabular}
    \end{centering}
\end{table}

The summary statistics in Figures~\ref{fig:cd_benchmark_covering} and~\ref{fig:cd_benchmark_ami} (bottom) and Table~\ref{tab:benchmark_summary} confirm the rankings. On average, CLaP scores $72.9 \pm 21.6$ / $60.4 \pm 30.9$ for Covering / AMI, representing an increase of $4.5$ / $16.1$ pp over the second-best competitor, ClaSP2Feat / Time2State. The differences in median are even more pronounced. Across both measures and all benchmarks, CLaP achieves the highest scores, except for SKAB, which aligns with the rankings. 

% This analysis demonstrates that CLaP is more accurate on average than its counterparts across the $405$ TS from the five benchmarks.

\subsection{Runtime and Scalability} \label{sec:runtime}

We measured the runtime of CLaP and its six competitors on the 405 TS to determine the time needed for TSSD, the accuracy-runtime tradeoff, as well as the scalability of CLaP.

\begin{figure}[t]
	\begin{minipage}{4cm}
        \includegraphics[width=1.0\columnwidth]{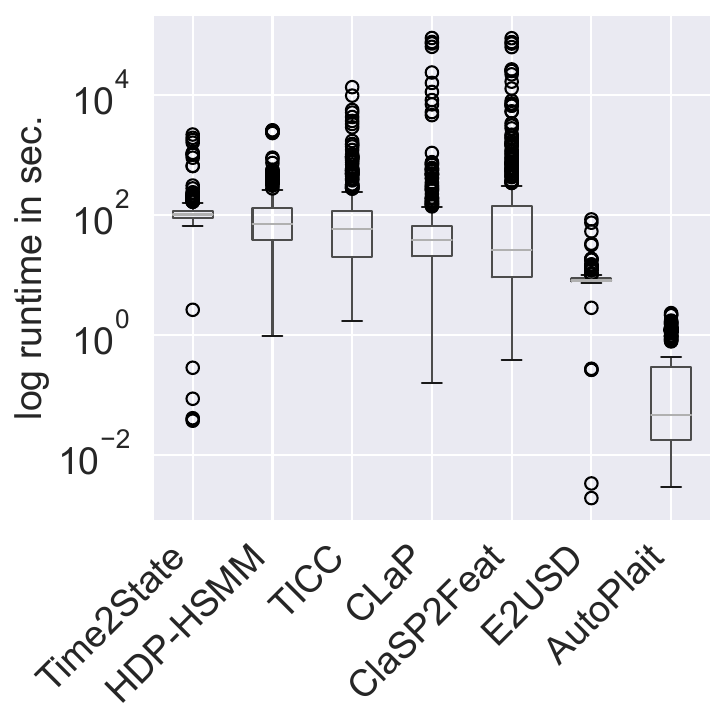}
	\end{minipage}
    \begin{minipage}{4cm}
        \includegraphics[width=1.0\columnwidth]{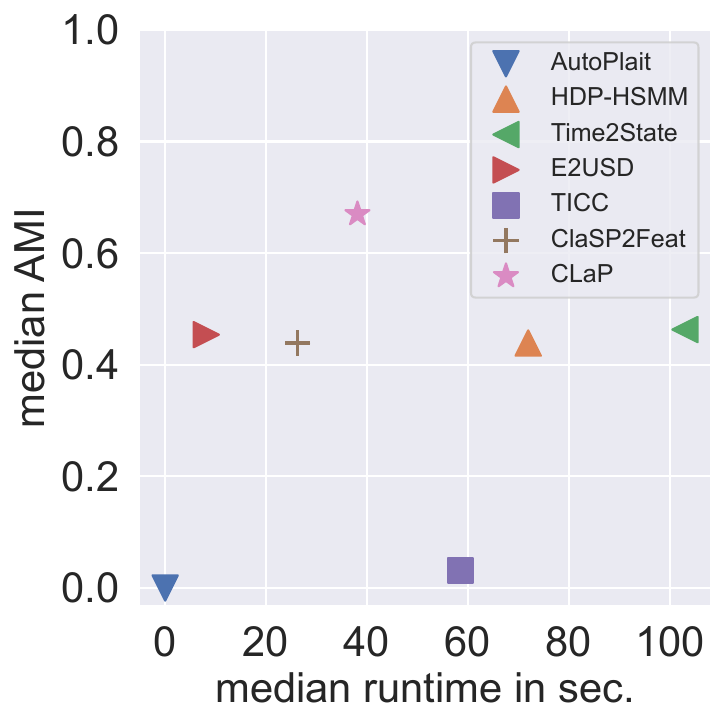}
	\end{minipage}
	\caption{Runtime distribution (left) and accuracy-runtime tradeoff (right) on $405$ TS for CLaP and six competitors.\label{fig:runtime}
	}
\end{figure}

\textbf{Runtime:} Figure~\ref{fig:runtime} (left) shows the runtime distributions of the methods on all data sets. AutoPlait (median of 0.05 seconds) leads with a substantial advantage over all other competitors due to its linear runtime complexity (with respect to TS size). Additionally, the main part of its implementation uses fast C code, while all other methods are implemented in Python. E2USD (8 seconds) and ClaSP2Feat (26 seconds) rank second and third. CLaP (38 seconds) secures fourth place. Its segmentation step takes 25 seconds, while the self-supervised classification and confused merging require 13 seconds. TICC (59 seconds) and HDP-HSMM (72 seconds) rank fifth and sixth, followed by Time2State (103 seconds) at last place. The distributions of all methods scatter widely, due to the large spans in TS lengths. CLaP is faster in median than three of its six competitors, but trades more computational costs, than e.g. AutoPlait, for higher accuracy. If runtime is a critical concern, users may choose to use the fast streaming implementation of ClaSP~\cite{Ermshaus2024ClaSS} or an even faster classifier, e.g. QUANT~\cite{Dempster2023QUANTAM}, which may, however, decrease accuracy.

\textbf{Accuracy vs. Runtime:} We investigated the tradeoff between accuracy and runtime of the approaches. Figure~\ref{fig:runtime} (right) depicts the median runtimes vs. AMI scores. The optimal solution is a fast and accurate algorithm (top left), while a poor algorithm is slow and inaccurate (bottom right). CLaP (pink star), ClaSP2Feat (brown plus), and E2USD (red triangle) all exhibit traits of fast and accurate algorithms. However, CLaP stands out with a solution that is only 12 seconds slower but $21.6$ pp more accurate. It offers, by far, the most desirable tradeoff between accuracy and runtime compared to the competitors on the considered TS.

\begin{figure}[t]
	\begin{minipage}{4cm}
        \includegraphics[width=1.0\columnwidth]{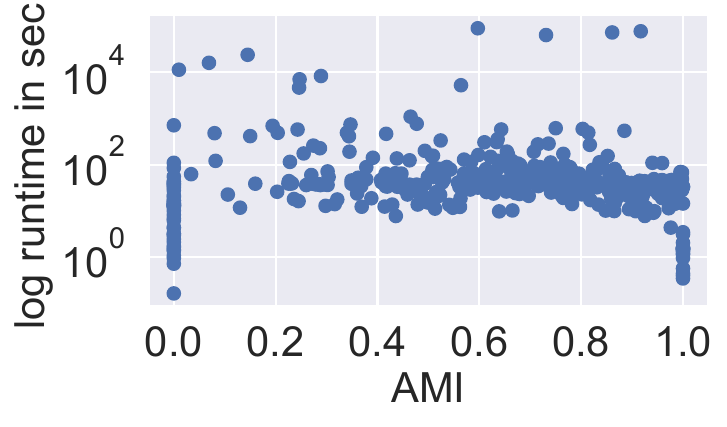}
	\end{minipage}
    \begin{minipage}{4cm}
        \includegraphics[width=1.0\columnwidth]{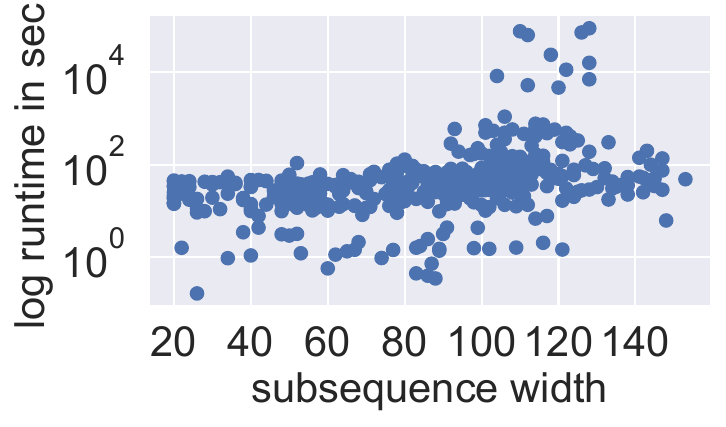}
	\end{minipage}
     \begin{minipage}{4cm}
        \includegraphics[width=1.0\columnwidth]{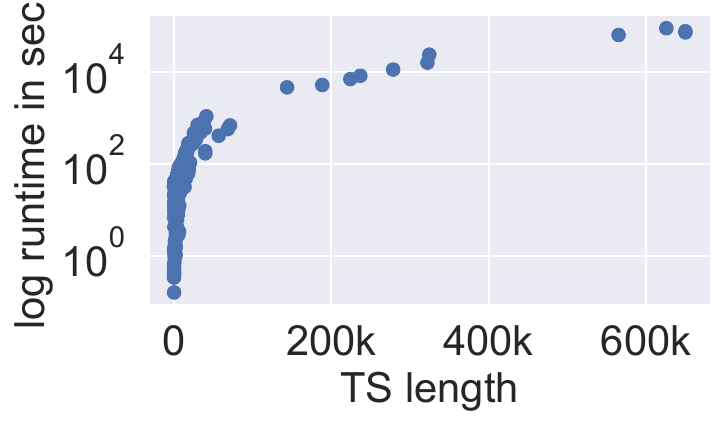}
	\end{minipage}
     \begin{minipage}{4cm}
        \includegraphics[width=1.0\columnwidth]{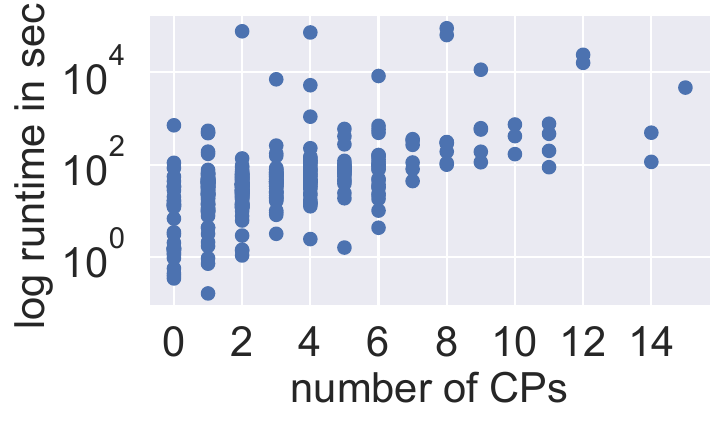}
	\end{minipage}
	\caption{Scalability of CLaP considering AMI performance (top left), subsequence width (top right), TS length (bottom left), and number of CPs (bottom right).\label{fig:scalability}
	}
\end{figure}

\textbf{Scalability:} Figure~\ref{fig:scalability} illustrates how CLaP's (log) runtime scales concerning four different variables, namely: AMI score, subsequence width, TS length, and number of CPs. For the first two variables, we do not observe clear relationships. As expected, CLaP becomes slower with increasing TS length. For instance, it requires less than 5 minutes to process up to 20k data points and less than 3 hours for up to 200k values. Its quadratic runtime complexity with respect to TS length can be roughly observed in the experimental runtimes. For an increasing number of CPs, CLaP also requires more runtime. However, the values are more dispersed compared to TS length. In summary, CLaP produces results for TS with tens of thousands of data points in minutes. For much larger data, runtime may become an issue, as for all other methods, except AutoPlait. In future work, we will investigate TS partitioning and hierarchical merging to further improve scalability.

\subsection{Ablation Study} \label{sec:ablation-study}

CLaP has four main components, namely: (a) a window size selection algorithm, (b) a segmentation procedure, (c) a classifier, and (d) a merge mechanism. We evaluated different approaches for each component, fixing the others to defaults, on $20\%$ of randomly chosen TS (21 out of 107) from TSSB and UTSA. To prevent overfitting, we excluded the remaining 86 out of 107 data sets, as well as the 298 TS from HAS, SKAB, and MIT-BIH from the ablations. The following analysis reports the summarized results. See the box plots for visualization and more evaluations on our website~\cite{CLaPWebpage}.

\textbf{(a) Window Size Selection:} To determine the window size for a TS in CLaP, we tested two whole-series-based methods: the most dominant Fourier frequency (FFT) and the highest autocorrelation offset (ACF) from~\cite{Ermshaus2022WSS}; as well as two subsequence-based algorithms: Multi-Window-Finder (MWF)~\cite{ImaniMultiWindowFinderDA} and Summary Statistics Subsequence (SuSS)~\cite{Ermshaus2022ClaSP}. Our experiments show no significant differences in average rank between the methods, which aligns with the results from~\cite{Ermshaus2022WSS}. We choose SuSS for WSS in CLaP, as it achieves the highest average scores for Covering of $80.5\% \pm 17.4\%$ and AMI of $76.1\% \pm 27.5\%$.

\textbf{(b) Segmentation Procedure:} We evaluated three optimization-based approaches for segmentation, namely BinSeg~\cite{Truong2020SelectiveRO}, PELT~\cite{Killick2011OptimalDO}, and Window~\cite{Truong2020SelectiveRO}, as well as the density-based approach FLUSS~\cite{Gharghabi2018DomainAO}, the density ratio estimator RuLSIF~\cite{Hushchyn2021Generalization}, and the self-supervised algorithm ClaSP~\cite{Ermshaus2022ClaSP}. For both metrics, ClaSP leads in average rank, with an insignificant advance for Covering (2.12) and a significant one for AMI (1.98). Hence, we use ClaSP in CLaP.

\iffalse
\begin{figure}[t]
	\begin{minipage}{4cm}
        \includegraphics[width=0.9\columnwidth]{figures/bp_ablation_study_window_size_train.pdf}
	\end{minipage}
	\begin{minipage}{4cm}
        \includegraphics[width=1.0\columnwidth]{figures/bp_ablation_study_segmentation_train.pdf}
	\end{minipage}
	\caption{AMI statistics for (a) window size selection (left), and (b) segmentation technique (right).\label{fig:bp_ablation_study_1}
	}
\end{figure}

\begin{figure}[t]
 	\begin{minipage}{4cm}
        \includegraphics[width=1.0\columnwidth]{figures/bp_ablation_study_classifier_train.pdf}
	\end{minipage}
	\begin{minipage}{4cm}
        \includegraphics[width=0.9\columnwidth]{figures/bp_ablation_study_merge_score_train.pdf}
	\end{minipage}
	\caption{AMI box plots for (c) classification algorithm (left), and (d) merge criterion (right).\label{fig:bp_ablation_study_2}
	}
\end{figure}
\fi

\textbf{(c) Classifier:} For self-supervised classification in CLaP, we tested six state-of-the-art classifiers according to~\cite{Middlehurst2023BakeOR}, namely: the feature-based approach FreshPRINCE~\cite{Middlehurst2022TheFA}, the interval-based algorithm QUANT~\cite{Dempster2023QUANTAM}, the shapelet-based method RDST~\cite{Guillaume2021RandomDS}, the dictionary-based procedure WEASEL 2.0~\cite{Schfer2023WEASEL2A}, and two convolution-based techniques, namely ROCKET~\cite{Dempster2019ROCKETEF} and MR-Hydra~\cite{Dempster2022HydraCC}. Our experiments show only differences in tendency between the average ranks of the classifiers. We choose ROCKET, as it has the highest-scoring median Covering of $81.6\%$ and AMI of $87.2\%$.

\textbf{(d) Merge Score:} To implement the merge mechanism, we evaluated four scores: our proposed cgain, F1, ROC/AUC, and AMI; we also assessed two loss functions: log and Hamming. Classification gain leads for both metrics in average rank, with an insignificant advance for Covering but a significant difference to unnormalized F1 and both losses in AMI. It achieves an increase in Covering (AMI) of $2.2$ ($4.1$) percentage points. Therefore, we use cgain in CLaP for confused merging.

In summary, the choice of window size and classification algorithm results in only negligible differences in performance, whereas the segmentation procedure and merge score have a substantial impact. This may be partially due to the interconnections between the components; for example, imprecisely located CPs can negatively influence self-supervised classification, and improper merging can easily degrade the entire state detection process.

We investigated whether CLaP’s two phases, segmentation and state identification, substantially contribute to TSSD by comparing each phase with a random baseline while keeping all other settings fixed. For segmentation, we ran CLaP with uniformly drawn CPs (without replacement), which lowered the average Covering score by $20.2$ pp and AMI by $31.8$ pp. For state identification, we assigned uniformly drawn labels to the segments (with replacement), causing Covering to drop by $20.5$ pp and AMI by $40.9$ pp.
These pronounced declines show that both phases are essential to CLaP’s performance; omitting either one sharply degrades accuracy.

\begin{figure}[t]
 	\begin{minipage}{4cm}
        \includegraphics[width=1.0\columnwidth]{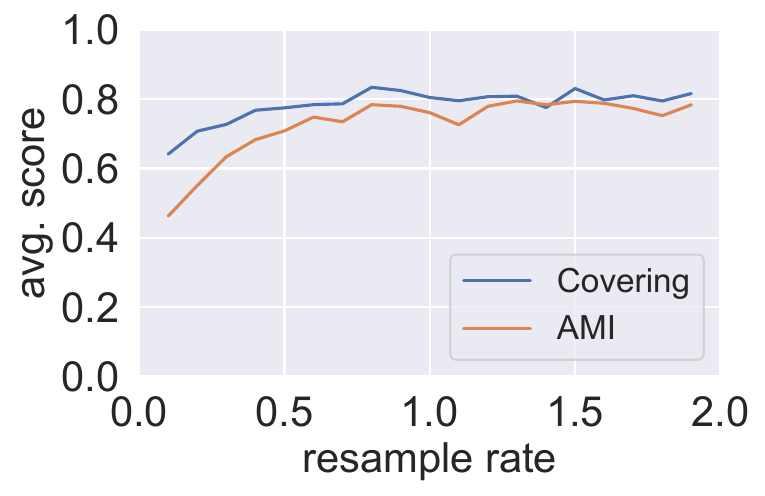}
	\end{minipage}
	\begin{minipage}{4cm}
        \includegraphics[width=1.0\columnwidth]{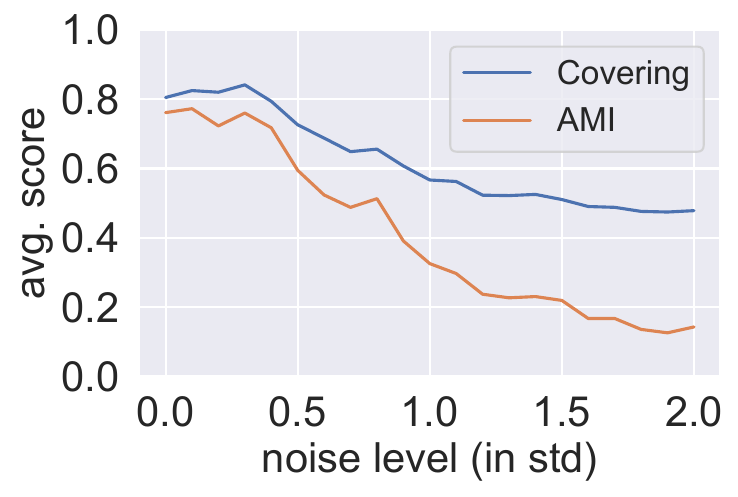}
	\end{minipage}
	\caption{Sensitivity analysis of CLaP regarding sampling rate (left) and noise level (right).\label{fig:sensitivity}
	}
\end{figure}

\subsection{Sensitivity Analysis}\label{sec:sensitivity}

Real-world TS data are collected at different sampling rates and often contain noise introduced by sensor artefacts or external events. We evaluated how these factors affect CLaP’s performance on the data sets from the ablation study to assess its robustness to changes in data representation.

\textbf{Sampling Rate:} Figure~\ref{fig:sensitivity} (left) plots CLaP’s average Covering and AMI scores when the data sets are resampled to 10\% up to 190\% (in 10 pp steps) of their original size. Both metrics increase steadily from 10\% to roughly 50\% and then plateau. This suggests that CLaP requires a minimum temporal resolution (30\% to 50\% for these data sets) to identify process states accurately, but maintains stable performance at higher resolutions. 

\textbf{Noise Level:} Figure~\ref{fig:sensitivity} (right) shows how CLaP’s performance changes when white noise with increasing standard deviation is added to the standardised data. Performance remains quite stable up to a noise level of 0.5, after which it declines sharply. Thus, CLaP is resilient to moderate noise but struggles to differentiate process states under very high noise levels.

These experiments show that CLaP is robust to variations in sampling rate and to moderate noise. When data become excessively coarse or noisy, performance degrades gradually, indicating that the method's performance scales with data quality.

\subsection{Discussion} \label{sec:discussion}

CLaP's central strength is its high accuracy. The comparative analysis shows that it is significantly more accurate compared to six state-of-the-art competitors on the five benchmarks. We carefully designed this edge by splitting TSSD into a two-step process: segmentation and state identification; optimizing each part separately:

\textbf{(a) Segmentation:} We use ClaSP for segmentation, which is one of the most accurate TSS algorithms for univariate and multivariate~\cite{Ermshaus2024MultivariateClaSP} batch data as well as streaming TS~\cite{Ermshaus2024ClaSS}. Except for ClaSP2Feat, the other competitors either do not use a dedicated segmentation procedure (e.g. Time2State) or use one with inferior detection accuracy, such as AutoPlait.

\textbf{(b) State Identification:} Our proposed self-supervised classification and confused merging procedure leverages the predictive power of classification algorithms~\cite{Middlehurst2023BakeOR}, which are more accurate than classical clustering strategies that do not use TS-specific feature engineering or label information, as i.e. implemented in TICC. Others rely on distance calculations~\cite{Holder2022ARA}, or use limited statistical models with assumptions (e.g. Time2State).

Both components substantially contribute to the overall performance, as the ablation study indicates for (a) and CLaP's advance over ClaSP2Feat shows for (b). Despite its strengths, CLaP has several weaknesses. First, it assumes that subsequences can fully capture the distinctive properties of process states. This assumption holds for TS with homogeneous segments that contain similar temporal patterns, but performance degrades when intra-state diversity is high, e.g. in the presence of high noise (see Figure \ref{fig:sensitivity}, right) or concept drift. Second, CLaP’s runtime is quadratic in both the TS length and the number of CPs. For very long sequences with many CPs, this leads to high latency and limits its direct use in streaming contexts. Although a streaming version of ClaSP exists~\cite{Ermshaus2024ClaSS}, further work is required to make confused merging viable for real-time processing. Consequently, the current CLaP implementation is best suited to batch TS with homogeneous segments.

% We explore the impact of these design choices in a real-world use case in Subsection~\ref{sec:usecases}, comparing them to the competitors.

\subsection{Satellite Image Data Use Case} \label{sec:usecases}

% We discuss the state detection results of CLaP and the second-best competitors, ClaSP2Feat and Time2State, to showcase their characteristics based on two particular TS. Both are part of the benchmark data sets.

\begin{figure}[t]
	\begin{minipage}{8cm}
        \includegraphics[width=1.0\columnwidth]{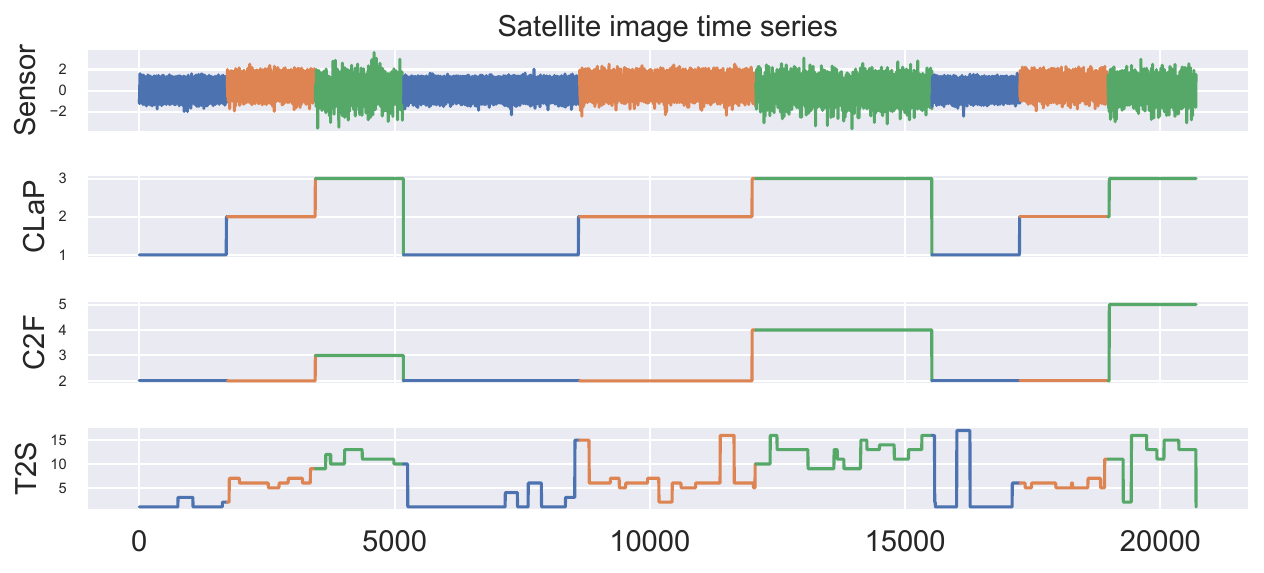}
	\end{minipage}
	\begin{minipage}{4cm}
        \includegraphics[width=0.8\columnwidth]{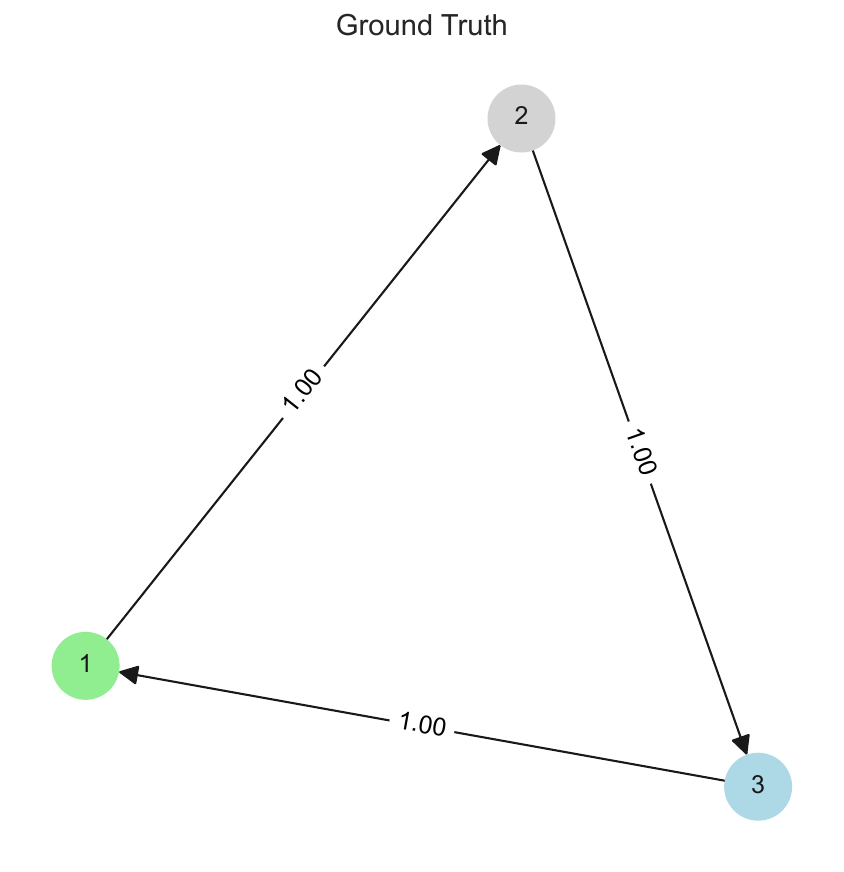}
	\end{minipage}
    \begin{minipage}{4cm}
        \includegraphics[width=0.8\columnwidth]{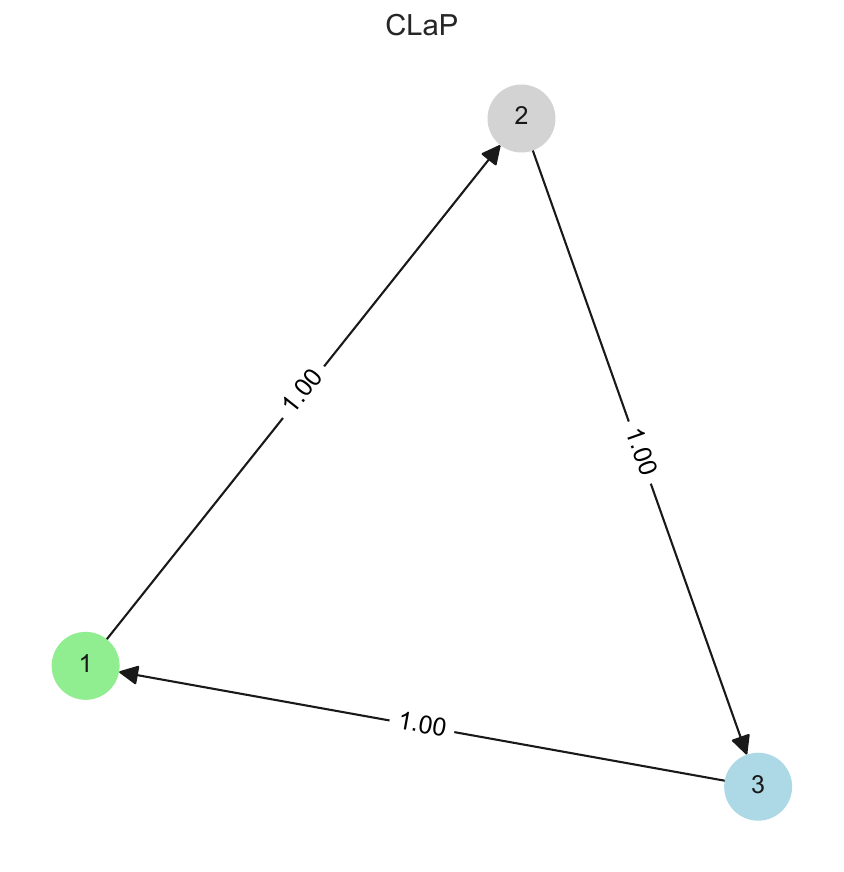}
	\end{minipage}
     \begin{minipage}{4cm}
        \includegraphics[width=0.8\columnwidth]{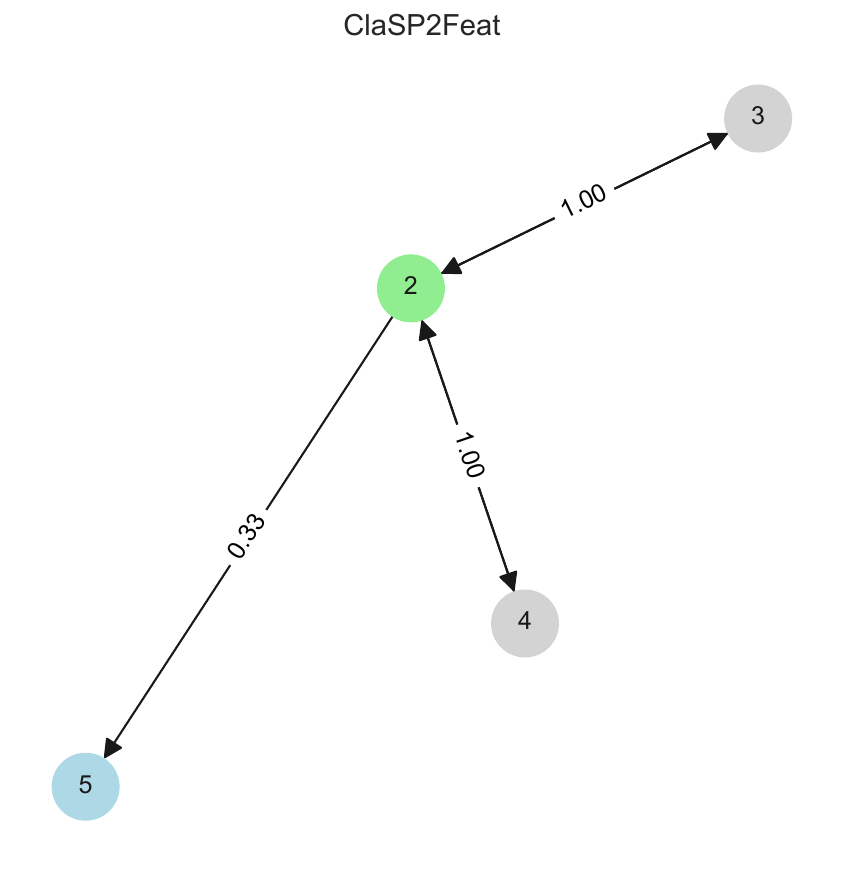}
	\end{minipage}
     \begin{minipage}{4cm}
        \includegraphics[width=0.8\columnwidth]{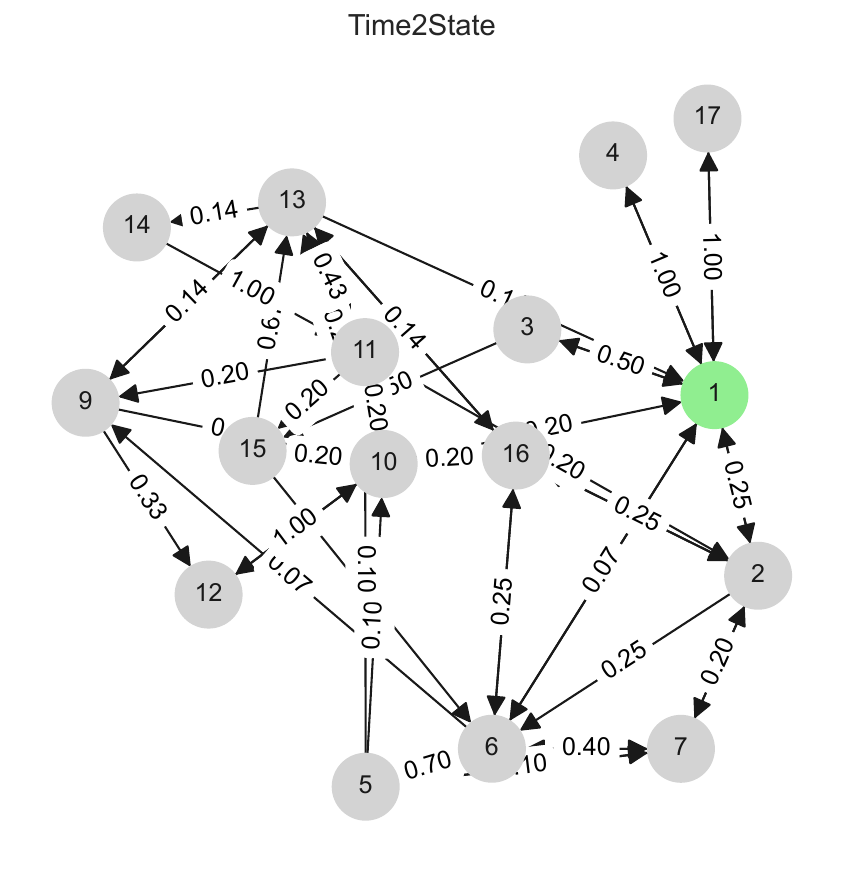}
	\end{minipage}
	\caption{TS (top) shows satellite image data for 3 different crops (blue, orange, green)~\cite{Schfer2021ClaSPT}. State sequences (2nd from top to fourth) shows results of CLaP, ClaSP2Feat, and Time2State. Bottom diagrams show graphical representations of ground truth and competitor results.\label{fig:satellite_image_usecase}
    }
\end{figure}

We revisit the satellite image example from Subsection~\ref{sec:classification-gain} in Figure~\ref{fig:satellite_image_usecase}. The TS (top) illustrates sensor data for 3 different crops (coloured in blue, orange, and green) across 9 segments. We computed the state sequences of CLaP, ClaSP2Feat, and Time2State (2nd from top to fourth). A well-fitting output annotates one distinct number to each observed state (e.g., 1 for blue, 2 for orange, and 3 for green). CLaP accurately identifies the boundaries of each segment and assigns correct state labels, resulting in the sequence: 1,2,3,1,2,3,1,2,3. ClaSP2Feat does not differentiate between the blue and orange segments and erroneously assigns different labels to the green segments. However, it correctly identifies some CPs. The result of Time2State is noisy and largely overestimates the total number of states. Nonetheless, it correctly labels many data points and identifies correct repetitions of segments. Figure~\ref{fig:satellite_image_usecase} (bottom) shows the state sequences as diagrams, with nodes representing states and edges depicting transitions. The start and end states appear in green (blue), and transition probabilities are labelled. It protrudes that only CLaP’s graph is isomorphic to the ground truth. This use case demonstrates the accuracy of CLaP for long and complex time series data (20.7k data points, 9 segments, 3 states). Its results are also interpretable for human inspection.

\section{Related Work} \label{sec:related-work}

The rapid growth of sensor data from IoT devices in \emph{smart} applications, such as healthcare and factories, has driven a substantial boost in research of TS management and mining~\cite{Krishnamurthi2020AnOO}. TS storage solutions include specialized databases~\cite{Pelkonen2015GorillaAF}, indices~\cite{Echihabi2022HerculesAD}, and compression algorithms~\cite{Liakos2022ChimpEL}. Unsupervised data analytics process the acquired data to automatically extract anomalies~\cite{Boniol2021UnsupervisedAS}, motifs~\cite{Schfer2022MotifletsS}, or clusters~\cite{Paparrizos2016kShapeEA}. This enables domain experts to make data-driven decisions, such as health professionals conducting gait analysis~\cite{Zhou2023DUOGAITAG}.

TSSD is a complex preprocessing step in a typical TS analysis workflow. Its main components, TS segmentation and clustering, have been researched both independently and in combination~\cite{Wang2024UnsupervisedTS}. TSS can be formalized as an optimization problem, where each segment incurs a cost based on a discrepancy measure (e.g., mean-shift~\cite{Page1955ATF} or entropy~\cite{Sadri2017Information}), and the number of segments~\cite{Truong2020SelectiveRO} is penalized. Exact or approximate solvers, such as PELT~\cite{Killick2012OptimalDO} or Wild BinSeg~\cite{Fryzlewicz2014Wild}, then compute the optimal segmentation. Adams et al. model segments as probability distributions and use a recursive message-passing algorithm to infer the most recent CP~\cite{Adams2007BayesianOC}. This approach can be extended to detect short gradual changes~\cite{Draayer2021ReevaluatingTC}, an extension of the original TSS problem~\cite{Carpentier2024PatternbasedTS}. A central limitation of these techniques is their dependence on domain-specific parameters, such as assumed value distributions or fitting cost functions. To overcome this, Katser et al. propose ensembling strategies, which increase robustness and accuracy~\citep{Katser2021UnsupervisedOC}. Other domain-agnostic methods include FLUSS~\cite{Gharghabi2018DomainAO} and ClaSP~\cite{Ermshaus2022ClaSP}, which achieve top-ranking results on recent TSS benchmarks~\cite{Ermshaus2023TimeSS,ErmshausHumanAS} and are capable of processing multivariate and streaming data. FLUSS measures the density of similar subsequences in potential segments as an arc curve, from which it extracts local minima that represent CPs~\cite{Gharghabi2018DomainAO}. ClaSP frames TSS as a collection of self-supervised subsequence classification problems and reports the segmentation with the highest cross-validation performance~\cite{Ermshaus2022ClaSP}. We use ClaSP in CLaP because it is hyper-parameter-free and makes few assumptions about segments, such as being mutually dissimilar.

After segmentation, individual segments must be assigned correct state labels, which can be achieved by clustering, for instance, by extracting equal-sized subsequences from segments, where the individual data points define the feature space. Partition-based algorithms, such as Lloyd’s algorithm~\cite{Lloyd1982LeastSQ}, PAM~\cite{Kaufman1990PartitioningAM}, or k-Shape~\cite{Paparrizos2016kShapeEA}, randomly partition the data and iteratively refine it until some convergence criterion is met~\cite{Holder2022ARA}. Specifically for TS, the data is commonly assigned to clusters based on minimal distances to their centres, such as medoids. Holder et al. evaluated the accuracy of different distance measures and found MSM~\cite{Stefan2013TheMM} and TWE~\cite{Marteau2007TimeWE} to be good choices~\cite{Holder2022ARA}. Different classical clustering approaches include density-based methods, e.g., DBSCAN~\cite{Ester1996ADA} and OPTICS~\cite{Ankerst1999OPTICSOP}, as well as hierarchical approaches, such as BIRCH~\cite{Zhang1996BIRCHAE}. These can be applied to the raw TS segments. However, Bonifati et al. report in a recent study that extracting and selecting features from TS before clustering improves performance~\cite{Bonifati2023InterpretableCO}. Specifically, they compute statistics with tsfresh~\cite{Christ2018TimeSF} from the data and calculate similarity measures between sequences. Features without variance are filtered, and the remaining ones are clustered. We use this approach in our baseline ClaSP2Feat, which combines an accurate segmentation and clustering procedure and achieves top-ranking results. 

Besides the independent segmentation and clustering, the literature also contains integrated TSSD approaches~\cite{Wang2024UnsupervisedTS}. HMMs, for instance, can model single TS with a limited number of states and probabilities for initialization, transition, and output~\cite{Bishop2006Pattern}. By design, HMM states model single data points, not prolonged events. Matsubara et al. tackle this by grouping HMM states and learning transitions between groups, which reveals their CPs. AutoPlait implements this idea by iteratively refining the segmentation to optimize a cost function~\cite{Matsubara2014Autoplait}. HDP-HSMM uses a different approach: it incorporates explicit-duration semi-Markovian properties to model flexible HMM state durations and employs a hierarchical Dirichlet process prior to learn the number of states from the TS directly~\cite{Johnson2010TheHD}. Hallac et al. propose characterizing TS using a sliding window instead of modelling single data points. Each subsequence belongs to a cluster that is characterized by a correlation network. In TICC, the assignment of subsequences and the update of cluster parameters are iteratively refined using an expectation maximization algorithm~\cite{Hallac2017ToeplitzIC}. The recent method Time2State also builds a sliding window but uses it to train a deep learning encoder by minimizing (maximizing) distances of intra-state (inter-state) subsequences. The resulting embedding is clustered using DPGMM to assign subsequences to state labels and to automatically learn their amount~\cite{Wang2023Time2StateAU}. E2USD implements a similar deep-learning TSSD architecture, transforming each sliding window through compression, trend–seasonal decomposition and contrastive learning into a compact embedding that is subsequently clustered via DPGMM~\cite{Lai2024E2UsdEU}.

Our proposed CLaP also processes subsequences, but differs substantially from the existing methods. It is the first TSSD method that uses self-supervised learning~\cite{Hido2008UnsupervisedCA} to exploit the predictive power of TS classification algorithms, which are more accurate than the unsupervised models used by the aforementioned methods that rely on classical techniques sensitive to hyper-parameters or TS characteristics. CLaP merges confused classes as long as their cgain increases, which clusters segments automatically in label space rather than in feature space.

Apart from state detection in single TS, current research also investigates the relationships among dimensions in MTS and across different data sets. Wang et al. examine high-dimensional TS in which a subset of channels must be automatically selected for accurate TSSD. They propose ISSD, an algorithm that identifies a subset of TS dimensions with high-quality states, quantified by channel set completeness, and solves this selection in an optimization problem~\cite{Wang2025ISSDIS}. StaCo is another algorithm designed to measure overall, partial, and time-lagged correlations between states in heterogeneous TS. It produces a state correlation matrix, from which state affiliations across data sets can be derived~\cite{Wang2024DetectingSC}.

% , for instance to support root cause analysis.

\section{Conclusion} \label{sec:conclusion}

We proposed CLaP (Classification Label Profile), a new time series state detection algorithm that uses self-supervised classification techniques for segmentation and labelling. Our experimental evaluation shows that CLaP combines high-scoring design choices, leading to a statistically significant improvement in accuracy over six competitors on 405 univariate and multivariate TS from five benchmarks. CLaP also offers the best tradeoff between accuracy and runtime. It computes state labels for tens of thousands of data points in just a few minutes, providing a useful TS annotation that can be employed as input for advanced data analytics or as decision support for domain experts.

% It merges labels based on confusion and classification gain to discover the latent states of prolonged events in time series. 

Limitations of CLaP include its sensitivity to the segmentation provided by ClaSP. While confused merging can correct false-positive CPs, it cannot relocate imprecise ones. It also has a quadratic runtime complexity with respect to TS size and the number of CPs. Some TS applications, such as predictive maintenance, require data analysis solutions capable of processing streaming data.

In future work, we plan to explore sampling, batching, and hierarchical techniques to improve the scalability of CLaP. We want to explore the incorporation of domain knowledge to CLaP to guide its TSSD. Experts often have specific information from visual inspection or access to exogenous data which could increase the quality of state detection, e.g. by indicating some CPs or state labels of single measurements. We further aim to investigate state detection in TS collections to identify recurring states and transitions across instances, potentially revealing even more semantics of the data-generating process.

% \begin{acks}
% \end{acks}

\bibliographystyle{ACM-Reference-Format}
\bibliography{main}

% \appendix

\end{document}